\pdfoutput=1
\documentclass[10pt,twocolumn,letterpaper]{article}

\usepackage{cvpr}
\usepackage{times}
\usepackage{epsfig}
\usepackage{graphicx}
\usepackage{amsmath}
\usepackage{amssymb}
\usepackage[ruled,vlined,linesnumbered]{algorithm2e} 
\usepackage{multirow}
\usepackage{booktabs}

\usepackage[breaklinks=true,bookmarks=false]{hyperref}

\cvprfinalcopy 

\def\cvprPaperID{8151} 

\ifcvprfinal\fi
\begin{document}

\title{Cross-domain Object Detection through Coarse-to-Fine Feature Adaptation}

\author{
Yangtao Zheng$^{1,\,2,\,3}$ \hspace{3mm} Di Huang$^{1,\,2,\,3}$\thanks{corresponding author} \hspace{4mm} Songtao Liu$^{1,\,2,\,3}$ \hspace{3mm} Yunhong Wang$^{1,\,3}$ \vspace{0.1cm} \\
$^1$Beijing Advanced Innovation Center for Big Data and Brain Computing, Beihang University \\
$^2$State Key Laboratory of Software Development Environment, Beihang University \\
$^3$School of Computer Science and Engineering, Beihang University, Beijing 100191, China \\
{\tt\small \{ytzheng,dhuang,liusongtao,yhwang\}@buaa.edu.cn}
}

\maketitle
\thispagestyle{empty}


\begin{abstract}
   Recent years have witnessed great progress in deep learning based object detection. However, due to the domain shift problem, applying off-the-shelf detectors to an unseen domain leads to significant performance drop. To address such an issue, this paper proposes a novel coarse-to-fine feature adaptation approach to cross-domain object detection. At the coarse-grained stage, different from the rough image-level or instance-level feature alignment used in the literature, foreground regions are extracted by adopting the attention mechanism, and aligned according to their marginal distributions via multi-layer adversarial learning in the common feature space. At the fine-grained stage, we conduct conditional distribution alignment of foregrounds by minimizing the distance of global prototypes with the same category but from different domains. Thanks to this coarse-to-fine feature adaptation, domain knowledge in foreground regions can be effectively transferred. Extensive experiments are carried out in various cross-domain detection scenarios. The results are state-of-the-art, which demonstrate the broad applicability and effectiveness of the proposed approach.
\end{abstract}



\section{Introduction}

\vspace{-1mm}

In the past few years, Convolutional Neural Networks (CNN) based methods have significantly improved the accuracies of plenty of computer vision tasks~\cite{he2016deep, long2015fully, fasterrcnn}. These remarkable gains often rely on large-scale benchmarks, such as ImageNet~\cite{deng2009imagenet} and MS COCO~\cite{coco}. Due to a phenomenon known as domain shift or dataset bias~\cite{Torralba2011Unbiased}, current CNN models suffer from performance degradation when they are directly applied to novel scenes. In practice, we are able to alleviate such an impact by  building a task-specific dataset that covers sufficiently  diverse samples. Unfortunately, it is rather expensive and time-consuming to annotate a large number of high-quality ground truths.

To address this dilemma, one promising way is to introduce Unsupervised Domain Adaptation (UDA) to transfer essential knowledge from an off-the-shelf labeled domain (referred to as the \textit{source domain}) to a related unseen but unlabeled one (the \textit{target domain})~\cite{Pan2010A}. Recently, UDA methods have been greatly advanced by deep learning techniques, and they mostly focus on generating domain-invariant deep representation by reducing cross-domain discrepancy (\eg Maximum Mean Discrepancy~\cite{Gretton2008A} or $\mathcal{H}$-divergence~\cite{Ben2010A}), which have proved very competent at image classification~\cite{Long2014Transfer, UDAbyBP, saito2017asymmetric, Ding_2018_ECCV, Kang_2019_CVPR} and semantic segmentation~\cite{CYCADA, Luo_2019_CVPR, Vu_2019_CVPR, Chen_2019_ICCV}. Compared to them, object detection is more complex, which is required to locate and classify all instances of different objects within images; therefore, how to effectively adapt a detector is indeed a challenging issue.

\begin{figure}[t]
   \centering
   \includegraphics[width=0.99\linewidth]{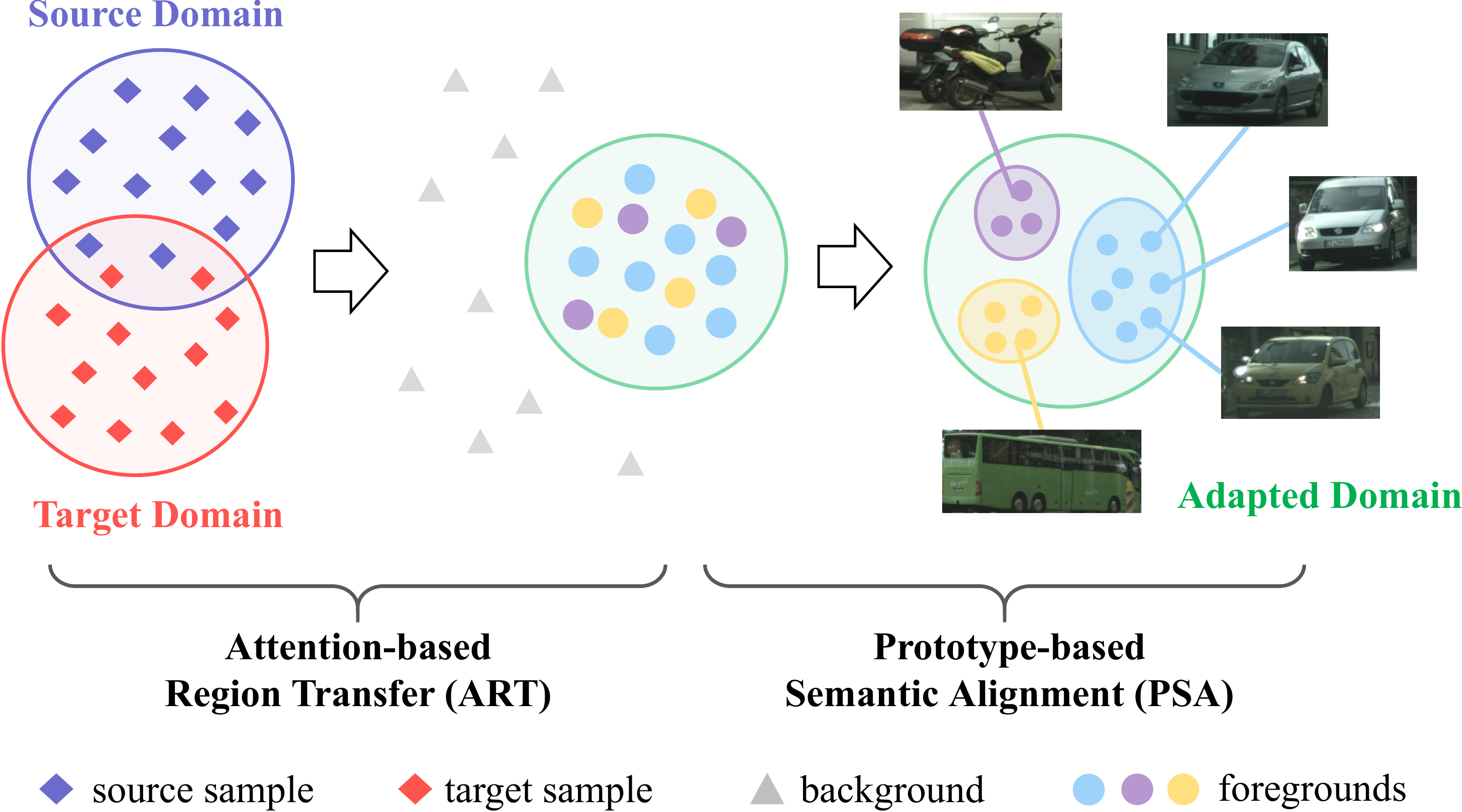}
   \vspace{1mm}
   \caption{Illustration of the proposed coarse-to-fine feature adaptation approach. It consists of two components, \ie, Attention-based Region Transfer (ART) and Prototype-based Semantic Alignment (PSA). The ART module figures out foregrounds from entire images in different domains and then aligns the marginal distributions on them. Further, the PSA module builds the prototype for each category to achieve semantic alignment. (Best viewed in color.)}
   \label{fig:brief} 
   \vspace{-2mm}
\end{figure}

In the literature, there are many solutions tackling this problem, including Semi-Supervised Learning (SSL) based \cite{Cai_2019_CVPR}, pixel-level adaptation based \cite{Kim_2019_CVPR, Hsu_2019_CVPR_Workshops, Adrian_2019_BMVC}, and feature-level adaptation based \cite{Chen_2018_CVPR,He_2019_ICCV, Xie_2019_ICCV_Workshops,Saito_2019_CVPR,zhu2019adapting}. The SSL based method reduces the domain gap through consistency regularization in a teacher-student scheme. However, the teacher does not always convey more meaningful knowledge than the student~\cite{Ke_2019_ICCV}, and the detector thus tends to accumulate errors, leading to deteriorated detection performance. Pixel-level based methods first conduct style transfer~\cite{zhu2017unpaired} to synthesize a target-like intermediate domain, aiming to limit visual shift, and then train detectors in a supervised manner. Nevertheless, it still remains a difficulty to guarantee the quality of generated images, especially in some extreme cases, which may hurt the adapted results. Alternatively, feature-level adaptation based methods mitigate the domain shift by aligning the features across domains. Such methods work more conveniently with competitive scores, making them dominate the existing community. 

In this category, domain adaptive Faster R-CNN~\cite{Chen_2018_CVPR} is a pioneer. It incorporates both image-level and instance-level feature adaptation into the detection model. In~\cite{Saito_2019_CVPR}, Strong-Weak feature adaptation is launched on image-level. This method mainly makes use of the focal loss to transfer hard-to-classify examples, since the knowledge in them is supposed to be more intrinsic for both domains. Although they deliver promising performance, image-level or instance-level feature adaptation is not so accurate as objects of interest locally distribute with diverse shapes. \cite{zhu2019adapting} introduces K-means clustering to mine transferable regions to optimize the adaptation quality. While attractive, this method highly depends on the pre-defined cluster number and the size of the grouped regions, which is not flexible, particularly to real-world applications. Furthermore, in the object detection task, there are generally multiple types of objects, and each has its own sample distribution. But these methods do not take such information into account and regard the distributions of different objects as a whole for adaptation, thereby leaving space for improvement.

In this paper, we present a coarse-to-fine feature adaptation framework for cross-domain object detection. The main idea is shown in Figure~\ref{fig:brief}. Firstly, considering that foregrounds between different domains share more common features compared to backgrounds~\cite{kim2019STABR}, we propose an Attention-based Region Transfer (ART) module to highlight the importance of foregrounds, which works in a class-agnostic coarse way. We extract foreground objects of interest by leveraging the attention mechanism in high-level features, and underline them during feature distribution alignment. Through multi-layer adversarial learning, effective domain confusion can be performed with the complex detection model. Secondly, category information of objects tends to further refine preceding feature adaptation, and in this case it is necessary to distinguish different kinds of foreground objects. Meanwhile, there is no guarantee that foregrounds of source and target images in the same batch have consistent categories, probably incurring object mis-matches in some mini-batch, making semantic alignment in UDA rather tough. Consequently, we propose a Prototype-based Semantic Alignment (PSA) module to build the global prototype for each category across domains. The prototypes are adaptively updated at each iteration, thus suppressing the negative influence of false pseudo-labels and class mis-matches. 

In summary, the contributions are three-fold as follows:
\vspace{-2mm}
\begin{itemize}
   \item A new coarse-to-fine adaptation approach is designed for cross-domain two-stage object detection, which progressively and accurately aligns deep features.
   \vspace{-2mm}
   \item Two adaptation modules, \ie, Attention-based Region Transfer (ART) and Prototype-based Semantic Alignment (PSA), are proposed to learn domain knowledge in foreground regions with category information.
   \vspace{-2mm}
   \item Extensive experiments are carried out in three major benchmarks in terms of some typical scenarios, and the results are state-of-the-art, demonstrating the effectiveness of the proposed approach.
\end{itemize} 



\section{Related Work}

\paragraph{Object Detection.} Object detection is a fundamental step in computer vision and has received increasing attention during decades. Most of traditional methods~\cite{VJ, HOG, DPM} depend on handcrafted features and sophisticated pipelines. In the era of deep learning, object detection can be mainly split into the one-stage detectors~\cite{redmon2016you, Liu2016SSD, lin2017focal, Liu_2018_ECCV} and the two-stage ones~\cite{girshick2014rich, girshick2015fast, fasterrcnn, lin2017feature}. However, those generic detectors do not address the domain shift problem that hurts detection performance in real-world scenes.
\vspace{-5mm}

\paragraph{Domain Adaptation.} Domain adaptation~\cite{ben2007analysis, Ben2010A} aims to boost performance in the target domain by leveraging common knowledge from the source domain, which has been widely studied in many visual tasks~\cite{xia2017detecting, Ding_2018_ECCV, Zhao_2019_CVPR, Luo_2019_ICCV, TAA_2019_ICCV, Fu_2019_ICCV}. With the advent of CNNs, many solutions reduce domain shift by learning domain-invariant features. Methods along this line can be divided into two streams: criterion-based~\cite{tzeng2014deep, long2015learning, sun2016deep} and adversarial learning-based~\cite{UDAbyBP, Tzeng_2017_CVPR, bousmalis2017unsupervised, pei2018multi}. The former aligns the domain distributions by minimizing some statistical distances between deep features, and the latter introduces the domain classifier to construct minimax optimization with the feature extractor. Despite great success is achieved, the majority of them can only handle relatively simple tasks, such as image classification.

\begin{figure*}[t]
   \centering
   \includegraphics[width=0.95\linewidth]{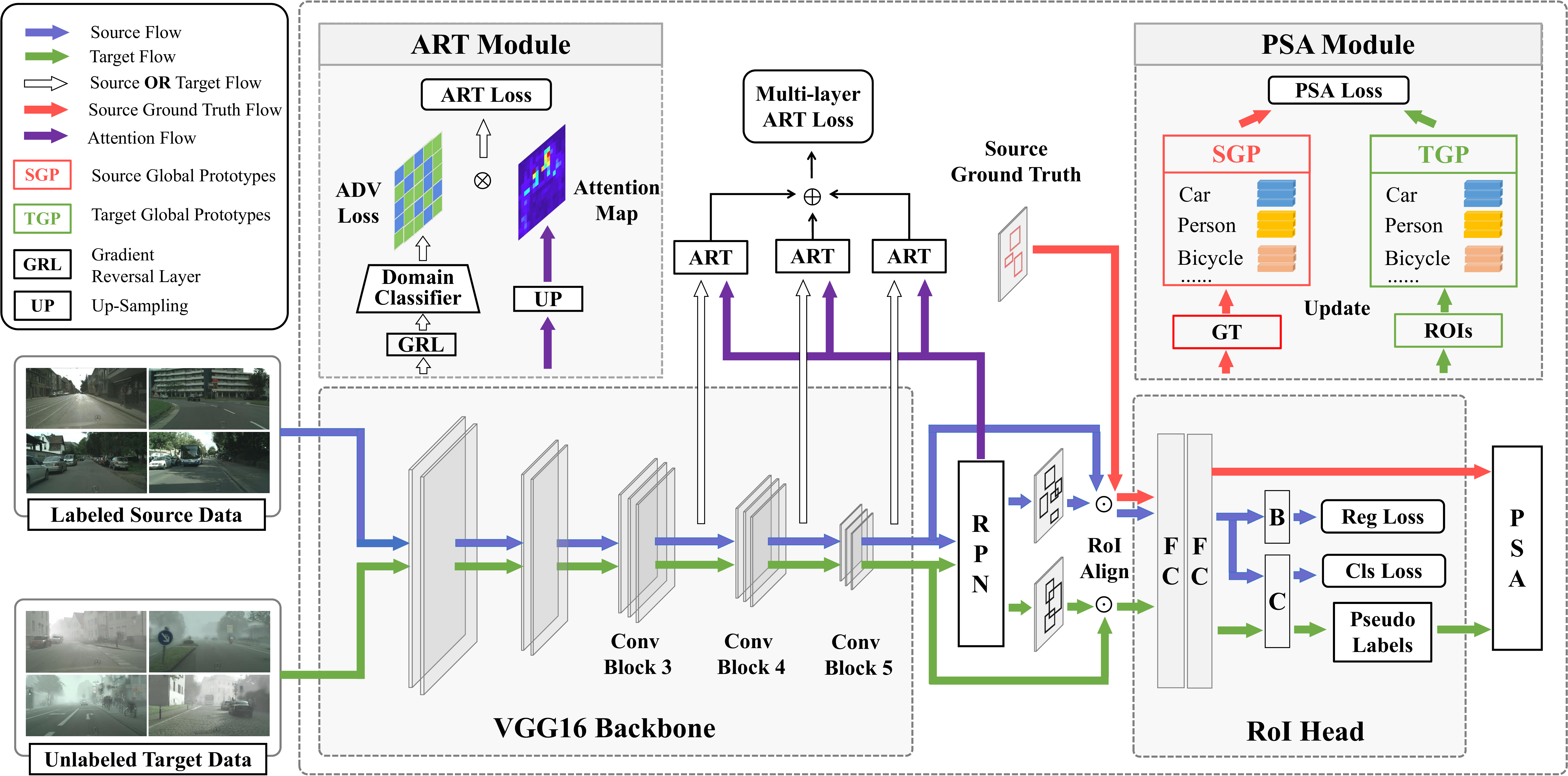}
   \caption{Overview of the proposed feature adaptation framework. We address the problem of domain shift on foreground regions by coarse-to-fine scheme with the ART and PSA modules. First, we utilize the attention map learned from the RPN module to localize foregrounds. Combined with multiple domain classifiers, the ART module puts more emphasis on aligning feature distributions of foreground regions, which achieves a coarse-grained adaptation in a category-agnostic way. Second, the PSA module makes use of ground truth labels (for source) and pseudo labels (for target) to maintain global prototypes for each category, and delivers fine-grained adaptation on foreground regions in a category-ware mode.}
   \label{fig:arch} 
   \vspace{-5mm}
\end{figure*}

\vspace{-5mm}
\paragraph{Cross-domain Object Detection.} A number of traditional studies~\cite{wang2011automatic, vazquez2012unsupervised, mirrashed2013domain, xu2014domain} focus on adapting a specific model (\eg, for pedestrian or vehicle detection) across domains. Later, \cite{raj2015subspace} proposes the adaptive R-CNN by subspace alignment~\cite{gopalan2011domain}. More Recently, the methods can be mainly grouped into four categories, including (1) \textit{Feature-level based}: \cite{Chen_2018_CVPR} presents domain adaptive Faster R-CNN to alleviate image-level and instance-level shifts, and \cite{He_2019_ICCV,Xie_2019_ICCV_Workshops} extend this idea to multi-layer feature adaptation. \cite{Saito_2019_CVPR} exploits strong-weak alignment components to attend strong matching in local features and weak matching in global features. \cite{zhu2019adapting} mines discriminative regions that contain objects of interest and aligns their features across domains. (2) \textit{SSL based}: \cite{Cai_2019_CVPR} integrates object relations into the measure of consistency cost with the mean teacher~\cite{tarvainen2017mean} model. (3) \textit{Pixel-level based}: \cite{Hsu_2019_CVPR_Workshops, Adrian_2019_BMVC} employ CycleGAN to translate the source domain to the target-like style. \cite{Kim_2019_CVPR} uses domain diversification and multi-domain invariant representation learning to address the imperfect translation and source-biased problem. (4) \textit{Others}: \cite{Khodabandeh_2019_ICCV} establishes a robust learning framework that formulates the cross-domain detection problem as training with noisy labels. \cite{kim2019STABR} introduces weak self-training and adversarial background score regularization for domain adaptive one-stage object detection. \cite{xu2019wasserstein} minimizes the wasserstein distance to improve the stability of adaptation. \cite{shen2019scl} explores a gradient detach based stacked complementary loss to adapt detectors.

As mentioned, feature-level adaptation is the main branch in cross-domain object detection, and its performance is currently limited by inaccurate feature alignment. The proposed method concentrates on two-stage detectors and substantially improves the quality of feature alignment by a coarse-to-fine scheme, where the ART module learns the adapted importance of foreground areas and the PSA module encodes the distribution property of each class. 

\vspace{-2mm}


\section{Method}
\vspace{-1mm}

\subsection{Problem Formulation}
\vspace{-1mm}

In the task of cross-domain object detection, we are given a labeled source domain $\mathcal{D}_S = \{ (x_i^s, y_i^s) \}_{i=1}^{N_s}$, where $x_i^s$ and $y_i^s=(b_i^s, c_i^s)$ denote the $i$-th image and its corresponding labels, \ie, the coordinates of the bounding box $b$ and its associated category $c$ respectively. In addition, we have access to an unlabeled target domain $\mathcal{D}_T = \{ x_i^t \}_{i=1}^{N_t}$. We assume that the source and target samples are drawn from different distributions (\ie, $\mathcal{D}_S \neq \mathcal{D}_T$) but the categories are exactly the same. The goal is to improve the detection performance in $\mathcal{D}_T$ using the knowledge in $\mathcal{D}_S$. 

\vspace{-1mm}
\subsection{Framework Overview}

\vspace{-1mm}
As shown in Figure~\ref{fig:arch}, we introduce a feature adaptation framework for cross-domain object detection, which contains a detection network and two adaptation modules.

\vspace{-3mm}
\paragraph{Detection Network.} We select the reputed and powerful Faster R-CNN~\cite{fasterrcnn} model as our base detector. Faster R-CNN is a two-stage detector that consists of three major components: 1) a backbone network $G$ that extracts image features, 2) a Region Proposal Network (RPN) that simultaneously predicts object bounds and objectness scores, and 3) a Region-of-Interest (RoI) head, including a bounding box regressor $B$ and a classifier $C$ for further refinement. The overall loss function of Faster R-CNN is defined as:
\begin{equation}
   \mathcal{L}_{det}(x) = \mathcal{L}_{rpn} + \mathcal{L}_{reg} + \mathcal{L}_{cls}
   \label{eq:sup}
\end{equation}
where $\mathcal{L}_{rpn}$, $\mathcal{L}_{reg}$, and $\mathcal{L}_{cls}$ are the loss functions for the RPN, RoI based regressor and classifier, respectively.

\vspace{-5mm}

\paragraph{Adaptation Modules.} Different from most of the existing studies which typically reduce domain shift in the entire feature space, we propose to conduct feature alignment on foregrounds that are supposed to share more common properties across domains. Meanwhile, in contrast to current methods that regard the samples of all objects as a whole, we argue that the category information contributes to this task and thus highlight the distribution of each category to further refine feature alignment. To this end, we design two adaptation modules, \ie, \textit{Attention-based Region Transfer} (ART) and \textit{Prototype-based Semantic Alignment} (PSA), to fulfill a coarse-to-fine knowledge transfer in foregrounds.

\subsection{Attention-based Region Transfer}

\vspace{-1mm}

The ART module is designed to raise more attention to align the distributions across two domains within the regions of foregrounds. It is composed of two parts: the domain classifiers and the attention mechanism. 

To align the feature distributions across domains, we integrate multiple domain classifiers $D$ into the last three convolution blocks in the backbone network $G$, where a two-player minimax game is constructed. Specifically, the domain classifiers $D$ try to distinguish which domain the features come from, while the backbone network $G$ aims to confuse the classifiers. In practice, $G$ and $D$ are connected by the Gradient Reverse Layer (GRL)~\cite{UDAbyBP}, which reverses the gradients that flow through $G$. When the training process converges, $G$ tends to extract domain-invariant feature representation. Formally, the objective of adversarial learning in the $l$-th convolution block can be written as:
\begin{equation}
   \begin{split}
      \mathcal{L}_{ADV}^{l} = \min \limits_{\theta_{G_l}} \max \limits_{\theta_{D_l}} \ & \mathbb{E}_{x_s \sim \mathcal{D}_S} \log D_l(G_l(x_s)) \\ 
       + \ & \mathbb{E}_{x_t \sim \mathcal{D}_T} \log (1 - D_l(G_l(x_t)))
   \end{split}
\end{equation}
where $\theta_{G_l}$ and $\theta_{D_l}$ are the parameters of $G_l$ and $D_l$ respectively. $D_l(\cdot)^{(h, w)}$ stands for the probability of the feature in location $(h, w)$ from the source domain.

\begin{figure}[t]
   \centering
   \includegraphics[width=0.9\linewidth]{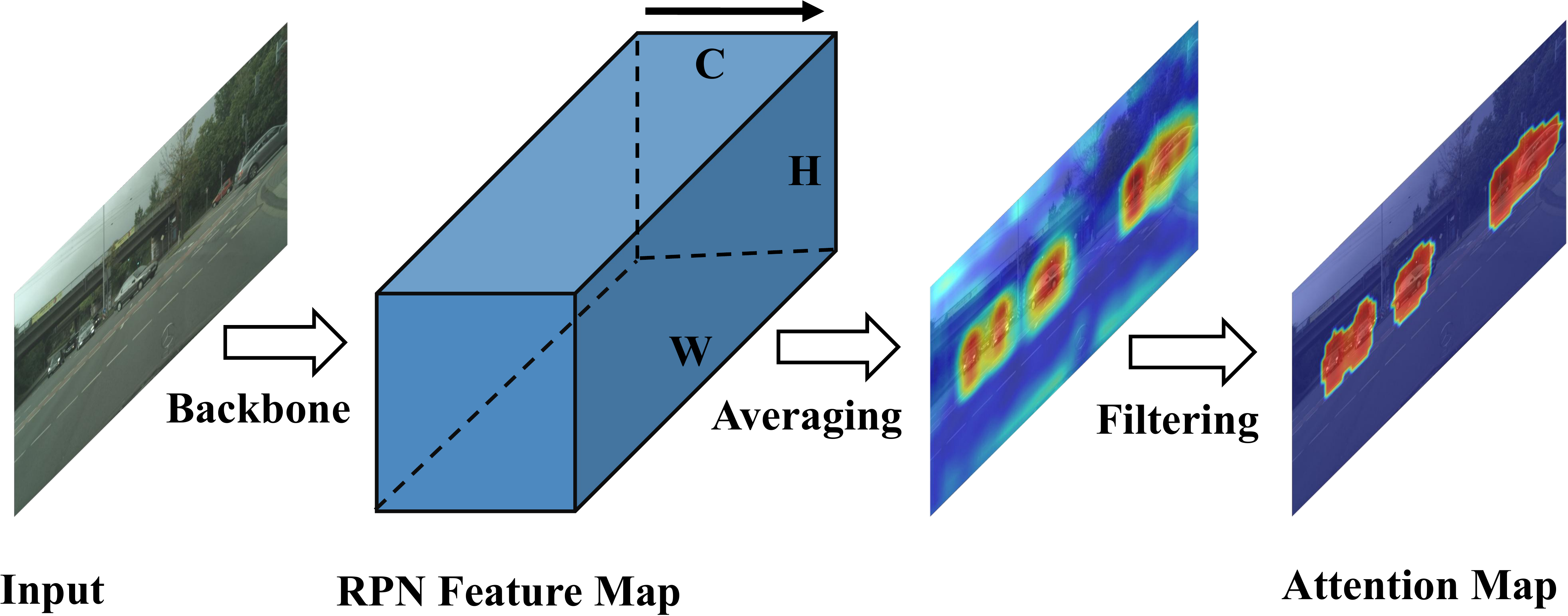}
   \caption{Illustration of the attention mechanism. We first extract the feature map from the RPN module. Then, we construct a spatial attention map by averaging values across the channel dimension. At last, filtering is applied to suppress the noise.}
   \label{fig:attention} 
   \vspace{-3mm}
\end{figure}

Recall that the detection task is required to localize and classify objects, and RoIs are usually more important than backgrounds. However, the domain classifiers align all the spatial locations of the whole image without focus, which probably degrades adaptation performance. To deal with this problem, we propose an attention mechanism to achieve foreground-aware distribution alignment. As mentioned in~\cite{fasterrcnn}, the RPN in Faster R-CNN serves as the attention to tell the detection model where to look, and we naturally utilize the high-level feature in RPN to generate the attention map, as shown in Figure~\ref{fig:attention}. To be specific, given an image $x$ from an arbitrary domain, we denote $F_{rpn}(x) \in \mathbb{R}^{H \times W \times C}$ as the output feature map of the convolutional layer in the RPN module, where $H \times W$ and $C$ are the spatial dimensions and the number of channels of the feature map, respectively. Then, we construct a spatial attention map by averaging activation values across the channel dimension. Further, we filter out (set to zero) those values that are less than the given threshold, which are more likely to belong to the background regions. The attention map $A(x) \in \mathbb{R}^{H \times W}$ is formulated as:
\begin{equation}
   M(x) = S(\frac{1}{C}\sum \limits_{c} | F_{rpn}^c(x) | )
\end{equation}
\begin{equation}
   T(x) = \frac{1}{HW}\sum \limits_{h,w} M(x)^{(h, w)}
\end{equation}
\begin{equation}
   A(x) = I(M(x) > T(x)) \otimes M(x) 
   \label{eq:at}     
\end{equation}
where $M(x)$ stands for the attention map before filtering. $S(\cdot)$ is the sigmoid function and $I(\cdot)$ is the indication function. $F_{rpn}^c(x)$ represents the $c$-th channel of the feature map. $\otimes$ denotes the element-wise multiplication. Threshold $T(x)$ is set to the mean value of $M(x)$. 

As the size of the attention map is not compatible with the features in different convolution blocks, we adopt bilinear interpolation to perform up-sampling, thus producing the corresponding attention maps. Due to the fact that the attention map may not always be so accurate, if a foreground region is mistaken for background, its attention weight is set to zero and cannot contribute to adaptation. Inspired by the residual attention network in~\cite{wang2017residual}, we add a skip connection to the attention map to enhance its performance. 

The total objective of the ART module is defined as:
\begin{equation}
   \begin{split}
      \mathcal{L}_{ART}  = \sum \limits_{l,h,w} (1 + U_l(A(x)^{(h,w)})) \cdot \mathcal{L}^{l,h,w}_{ADV}
   \end{split}
   \label{eq:art}
\end{equation}
where $U_l(\cdot)$ is the up-sampling operation and $\mathcal{L}^{l,h,w}_{ADV}$ stands for the adversarial loss on pixel $(h, w)$ in the $l$-th convolution block. Combining adversarial learning with the attention mechanism, the ART module aligns the feature distributions of foreground regions that are more transferable for the detection task.

\subsection{Prototype-based Semantic Alignment}

\vspace{-1mm}

Since the attention map from RPN carries no information about classification, the ART module aligns the feature distributions of foregrounds in a category-agnostic way. To achieve class-aware semantic alignment, a straightforward method is to train domain classifiers for each category. Nevertheless, there are two main disadvantages: (1) training multiple class-specific classifiers is inefficient; (2) false pseudo-labels (\eg, backgrounds or misclassified foregrounds) occurred in the target domain may hurt the performance of semantic alignment. 

Inspired by the prototype-based methods in few-shot learning~\cite{snell2017prototypical} and cross-domain image classification~\cite{xie2018learning, chen2019progressive, pan2019transferrable}, we propose the PSA module to handle the above problems. Instead of directly training classifiers, PSA tries to minimize the distance between the pair of prototypes $(P^S_k, P^T_k)$ with the same category across domains, thus maintaining the semantic consistency in the feature space. Formally, the prototypes can be defined as:
\begin{equation}
   P^S_k = \frac{1}{|GT_k|} \sum \limits_{r \in GT_k} F(r)
   \label{eq:psk}
\end{equation}
\begin{equation}
   P^T_k = \frac{1}{|RoI_k|} \sum \limits_{r \in RoI_k} F(r)
   \label{eq:ptk}
\end{equation}
where $P^S_k$ and $P^T_k$ represent the prototypes of the $k$-th category in the source and target domain respectively. $F(r)$ denotes the feature of foreground region $r$ after the second fully-connected (FC) layer in the RoI head. We use the ground truth $GT_k$ to extract the foreground regions in the source domain. Due to the absence of target annotations, we employ the $RoI_k$ provided by the RoI head module as the pseudo labels in the target domain. $|\cdot|$ indicates the number of regions.

The benefits of prototypes are two-fold: (1) the prototypes have no extra trainable parameters and can be calculated in linear time; (2) the negative influence of false pseudo-labels can be suppressed by the correct ones whose number is much larger when generating the prototypes. It should be noted that the prototypes above are built over all samples. In the training process, the size of each mini-batch is usually small (\eg, 1 or 2) for the detection task, and the foreground objects of source and target images in the same batch may have inconsistent categories, making categorical alignment not practical for all classes at this batch. For example, two images (one for each domain) are randomly selected for training, but \textit{Car} only appears in the source image. As a consequence, we cannot align the prototypes of \textit{Car} across domains in this batch. 

\begin{algorithm}  
   \SetKwFunction{Sample}{Sample}
   \caption{The coarse-to-fine feature adaptation framework for cross-domain object detection.}    
   \label{alg:1}
   \KwIn {Labeled source domain $\mathcal{D}_S$.\\
   \qquad \quad Unlabeled target domain $\mathcal{D}_T$.\\
   \qquad \quad Batch size $B$. Category number $C$.\\ }
   \KwOut {An adaptive detector $F(\cdot;\theta)$.}
   {Calculate the initial global prototypes $GP^{S(0)}_{k}$ and $GP^{T(0)}_{k}$ using the pretrained detector based on $\mathcal{D}_S$\\}
   \For {$i = 1$ \KwTo $max\_iter$} {
      {$X_S$, $Y_S$ $\gets$ \Sample{$\mathcal{D}_S$, $B / 2$}\\}
      {$X_T$ $\gets$ \Sample{$\mathcal{D}_T$, $B / 2$}\\}
      {\textbf{Supervised Learning:}\\}
      {Calculate $\mathcal{L}_{det}$ according to Eq.~(\ref{eq:sup})\\}
      {\textbf{Coarse-grained Adaptation:}\\}
      {Calculate $A(X_S)$ and $A(X_T)$ by Eq.~(\ref{eq:at})\\}
      {Calculate $\mathcal{L}_{ART}$ by Eq.~(\ref{eq:art})\\}
      {\textbf{Fine-grained Adaptation:}\\}
      {$\hat{Y_T}$ $\gets$ $F(X_T; \theta)$\\}
      \For {$k = 1$ \KwTo $C$} {
         {Calculate $P^{S(i)}_k$ and $P^{T(i)}_k$ by Eq.~(\ref{eq:psk}) and (\ref{eq:ptk})\\}
         {Update $GP^{S(i)}_{k}$ and $GP^{T(i)}_{k}$ by Eq.~(\ref{eq:gp})\\} 
      }
      {Calculate $\mathcal{L}_{PSA}$ according to Eq.~(\ref{eq:psa})\\}
      {Optimize the detection model by Eq.~(\ref{eq:loss})\\}
   }
\end{algorithm} 

To tackle the problem, we dynamically maintain global prototypes, which are adaptively updated by local prototypes at each mini-batch as follows:
\begin{equation}
   \alpha = sim(P^{(i)}_{k}, GP^{(i-1)}_{k})
\end{equation}
\begin{equation}
   GP^{(i)}_{k} = \alpha P^{(i)}_{k} + (1 - \alpha) GP^{(i-1)}_{k}
   \label{eq:gp}
\end{equation}
where $sim(x_1, x_2) = (\frac{x^\mathsf{T}_1 \cdot x_2}{\|x_1\|\|x_2\|} + 1) / 2$ denotes the cosine similarity. $P^{(i)}_{k}$ represents the local prototypes of the $k$-th category at $i$-th iteration. It is worth noting that we calculate the initial global prototypes $GP^{(0)}_{k}$ by Eq.~(\ref{eq:psk}) (for source) and Eq.~(\ref{eq:ptk}) (for target) based on the pretrained model from the labeled source domain. 

We do not directly align the local prototypes, but minimize the $L_2$ distance between the source global prototypes $GP^{S}_{k}$ and the target global prototypes $GP^{T}_{k}$ to achieve semantic alignment. The objective of the PSA module at $i$-th iteration can be formulated as following:
\begin{equation}
   \mathcal{L}_{PSA} = \sum \limits_{k} \| GP^{S(i)}_{k} - GP^{T(i)}_{k} \|^2 
   \label{eq:psa}
\end{equation}

\begin{table*}[t]
    \begin{center}
    \scriptsize
    \begin{tabular}{c|c|cccccccc|c|c|c} 
    \toprule
    \multicolumn{13}{c}{\textbf{Cityscapes $\rightarrow$ FoggyCityscapes}} \\
    \midrule 
        Method & Arch. & Bus & Bicycle & Car & Motor & Person & Rider & Train & Truck & \textbf{mAP} &  \textbf{mAP*} &\textbf{Gain}\\
    \midrule 
    \midrule 
    MTOR \cite{Cai_2019_CVPR}               & R & 38.6 & 35.6 & 44.0 & 28.3 & 30.6 & 41.4 & 40.6 & 21.9 & 35.1 &  26.9 & 8.2 \\
    RLDA \cite{Khodabandeh_2019_ICCV}       & I & 45.3 & 36.0 & 49.2 & 26.9 & 35.1 & 42.2 & 27.0 & 30.0 & 36.5 &  31.9 & 4.6 \\
    \midrule \midrule

    DAF \cite{Chen_2018_CVPR}               & V & 35.3 & 27.1 & 40.5 & 20.0 & 25.0 & 31.0 & 20.2 & 22.1 & 27.6 &  18.8 & 8.8 \\ 
    SCDA \cite{zhu2019adapting}             & V & 39.0 & 33.6 & 48.5 & 28.0 & 33.5 & 38.0 & 23.3 & 26.5 & 33.8 &  26.2 & 7.6 \\
    MAF \cite{He_2019_ICCV}                 & V & 39.9 & 33.9 & 43.9 & 29.2 & 28.2 & 39.5 & \underline{33.3} & 23.8 & 34.0 &  18.8 & 15.2 \\
    SWDA \cite{Saito_2019_CVPR}             & V & 36.2 & 35.3 & 43.5 & 30.0 & 29.9 & 42.3 & 32.6 & 24.5 & 34.3 &  20.3 & 14.0 \\
    DD-MRL \cite{Kim_2019_CVPR}             & V & 38.4 & 32.2 & 44.3 & 28.4 & 30.8 & 40.5 & \textbf{34.5} & 27.2 & 34.6 &  17.9 & 16.7 \\
    MDA \cite{Xie_2019_ICCV_Workshops}      & V & 41.8 & 36.5 & 44.8 & 30.5 & 33.2 & 44.2 & 28.7 & 28.2 & 36.0 &  22.8 & 13.2 \\ 
    PDA \cite{Hsu_2019_CVPR_Workshops}      & V & \underline{44.1} & 35.9 & \textbf{54.4} & 29.1 & \textbf{36.0} & 45.5 & 25.8 & 24.3 & 36.9 &  19.6 & \underline{17.3} \\
    \midrule
    Source Only    & V & 25.0 & 26.8 & 30.6 & 15.5 & 24.1 & 29.4 & 4.6  & 10.6 & 20.8 & -                     & -    \\
    3DC (Baseline)       & V & 37.9 & 37.1 & 51.6 & 33.1 & 32.9 & 45.6 & 27.9 & 28.6 & 36.8 & 20.8 & 16.0 \\
    Ours w/o ART & V & 41.6 & 35.4 & 51.5 & \textbf{36.9} & 33.5 & 45.2 & 26.6 & 28.2 & 37.4 & 20.8 & 16.6 \\
    Ours w/o PSA & V & \textbf{45.2} & \underline{37.3} & 51.8 & 33.3 & 33.9 & \underline{46.7} & 25.5 & \underline{29.6} & \underline{37.9} & 20.8 & 17.1 \\
    Ours           & V &  43.2 & \textbf{37.4} & \underline{52.1} & \underline{34.7} & \underline{34.0} & \textbf{46.9} & 29.9 & \textbf{30.8} & \textbf{38.6} & 20.8 & \textbf{17.8} \\
    \midrule
    \midrule 
        Oracle         & V & 49.5&37.0&52.7&36.0&36.1&47.1&56.0&32.1&43.3&-&- \\
    \bottomrule
    \end{tabular}
    \end{center}
    \caption{
        Results (\%) of different methods in the Normal-to-Foggy adaptation scenario. ``V'', ``R'' and ``I'' represent the VGG16, ResNet50 and Inception-v2 backbones respectively. ``Source Only'' denotes the Faster R-CNN model trained on the source domain only. ``3DC'' stands for the Faster R-CNN model integrated with three domain classifiers, which is our baseline method. ``Oracle'' indicates the model trained on the labeled target domain.  \textbf{mAP*} shows the result of ``Source Only'' for each method, and \textbf{Gain} displays its the improvement after adaptation. The best results are \textbf{bolded} and the second best results are \underline{underlined} among the methods with the VGG16 backbone.
    }
    \label{table:city-foggy}
    \vspace{-5mm}
\end{table*}

\vspace{-5mm}

\subsection{Network Optimization}

The training procedure of our proposed framework integrates three major components, as shown in Algorithm~\ref{alg:1}.

\vspace{-2mm}
\begin{enumerate}
   \setlength{\itemsep}{3pt}
   \setlength{\parsep}{0pt}
   \setlength{\parskip}{0pt}
   \item \textbf{Supervised Learning.} The supervised detection loss $\mathcal{L}_{det}$ is only applied to the labeled source domain $\mathcal{D}_S$.
   \item \textbf{Coarse-grained Adaptation.} We utilize the attention mechanism to extract the foregrounds in images. Then, we focus on aligning the feature distributions of those regions by optimizing $\mathcal{L}_{ART}$.
   \item \textbf{Fine-grained Adaptation.} At first, pseudo labels are predicted in the target domain. We further update the global prototypes for each category adaptively. Finally, semantic alignment on foreground objects is achieved by optimizing $\mathcal{L}_{PSA}$.
\end{enumerate}
\vspace{-3mm}
With the terms aforementioned, the overall objective is:
\begin{equation}
   \mathcal{L}_{total} = \mathcal{L}_{det} + \lambda_{1} \mathcal{L}_{ART} + \lambda_{2} \mathcal{L}_{PSA},
   \label{eq:loss}
\end{equation}
where $\lambda_{1}$ and $\lambda_{2}$ denote the trade-off factors for the ART module and the PSA module, respectively. 
\vspace{-2mm}


\section{Experiments}
\vspace{-1mm}

\subsection{Datasets and Scenarios}

\vspace{-1mm}
\textbf{Datasets.} Four datasets are used in evaluation. (1) \textbf{\textit{Cityscapes}}~\cite{cordts2016cityscapes} is a benchmark for semantic urban scene understanding. It contains 2,975 training images and 500 validation images with pixel-level annotations. Since it is not designed for the detection task, we follow~\cite{Chen_2018_CVPR} to use the tightest rectangle of an instance segmentation mask as the ground truth bounding box. (2) \textbf{\textit{FoggyCityscapes}}~\cite{sakaridis2018semantic} derives from Cityscapes by adding synthetic fog to the original images. Thus, the train/val split and annotations are the same as those in Cityscapes. (3) \textbf{\textit{SIM10k}}~\cite{SIM10K} is a synthetic dataset containing 10,000 images, which is rendered from the video game Grand Theft Auto V (GTA5). (4) \textbf{\textit{KITTI}}~\cite{Geiger2012CVPR} is another popular dataset for autonomous driving. It consists of 7,481 labeled images for training. 

\textbf{Scenarios.} Following~\cite{Chen_2018_CVPR}, we evaluate the framework under three adaptation scenarios as follows:

(1) \textbf{\textit{Normal-to-Foggy}} (Cityscapes $\rightarrow$ FoggyCityscapes). It aims to perform adaptation across different weather conditions. During the training phase, we use the training set of \textit{Cityscapes} and \textit{FoggyCityscapes} as the source and target domain respectively. Results are reported in the validation set of \textit{FoggyCityscapes}.

(2) \textbf{\textit{Synthetic-to-Real}} (SIM10k $\rightarrow$ Cityscapes). Synthetic images offer an alternative to alleviate the data annotation problem. To adapt the synthetic scenes to the real one, we utilize the entire \textit{SIM10k} dataset as the source domain and the training set of \textit{Cityscapes} as the target domain. Since only \textit{Car} is annotated in both domains, we report the performance of \textit{Car} in the validation set of \textit{Cityscapes}.

(3) \textbf{\textit{Cross-Camera}} (Cityscapes $\rightarrow$ KITTI). Images captured by different devices or setups also incur the domain shift problem. To simulate this adaptation, we use the training set of \textit{Cityscapes} as the source domain and the training set of \textit{KITTI} as the target domain. Note that the classification standards of categories in the two domains are different, we follow~\cite{Xie_2019_ICCV_Workshops} to classify \{\textit{Car}, \textit{Van}\} as \textit{Car}, \{\textit{Pedestrian}, \textit{Person sitting}\} as \textit{Person}, \textit{Tram} as \textit{Train}, \textit{Cyclist} as \textit{Rider} in \textit{KITTI}. The results are reported in the training set of \textit{KITTI}, which is the same as in~\cite{Chen_2018_CVPR,Xie_2019_ICCV_Workshops}.

\vspace{-1mm}
\subsection{Implementation Details}
\vspace{-2mm}
In all experiments, we adopt the Faster R-CNN with the VGG16~\cite{vgg} backbone pre-trained on ImageNet~\cite{deng2009imagenet}. We resize the shorter sides of all images to 600 pixels. The batch size is set to 2, \ie, one image per domain. The detector is trained with SGD for 50k iterations with the learning rate of $10^{-3}$, and it is then dropped to $10^{-4}$ for another 20k iterations. Domain classifiers are trained by the Adam optimizer~\cite{kingma2014adam} with the learning rate of $10^{-5}$. The factor $\lambda_1$ is set at 1.0. Since prototypes in the target domain are unreliable at the beginning, the PSA module is employed after 50k iterations with $\lambda_2$ set at 0.01. We report mAP with an IoU threshold of 0.5 for evaluation.

\begin{table}[t]
    \scriptsize
    \begin{center}
    \setlength{\tabcolsep}{8pt}
    \begin{tabular}{c|c|c|c|c} 
    \toprule
    \multicolumn{5}{c}{\textbf{SIM10k $\rightarrow$ Cityscapes}} \\
    \midrule 
        Method & Arch. & \textbf{AP on Car} & \textbf{AP*} & \textbf{Gain}\\
    \midrule \midrule
    RLDA \cite{Khodabandeh_2019_ICCV}       & I & 42.6 & 31.1 &  11.5 \\ 
    MTOR \cite{Cai_2019_CVPR}               & R & 46.6 & 39.4 &  7.2 \\
    \midrule \midrule

    DAF \cite{Chen_2018_CVPR}               & V & 39.0 & 30.1 &  8.9 \\
    MAF \cite{He_2019_ICCV}                 & V & 41.1 & 30.1 &  \textbf{11.0} \\
    SWDA \cite{Saito_2019_CVPR}             & V & 42.3 & 34.6 &  7.7 \\
    MDA \cite{Xie_2019_ICCV_Workshops}      & V & 42.8 & 34.3 &  8.5 \\ 
    SCDA \cite{zhu2019adapting}             & V & 43.0 & 34.0 &  \underline{9.0} \\
    \midrule 

    Source Only & V & 35.0 & - & - \\
    3DC (Baseline)  & V & 42.3 & 35.0 & 7.3 \\
    Ours w/o ART  & V & 42.7 & 35.0 & 7.7 \\
    Ours w/o PSA  & V & \underline{43.4} & 35.0 & 8.4 \\
    Ours  & V & \textbf{43.8} & 35.0 & 8.8 \\

    \midrule \midrule
        Oracle & V & 59.9 & - & - \\
    \bottomrule
    \end{tabular}
    \end{center}
    \caption{
        Results (\%) of the Synthetic-to-Real adaptation scenario.
    }
    \label{table:sim-city}
    \vspace{-5mm}
\end{table}

\subsection{Results}
\vspace{-1mm}
We conduct extensive experiments and make comparison to the state-of-the-art cross-domain object detection methods, including (1) \textbf{Semi-Supervised Learning}: MTOR~\cite{Cai_2019_CVPR}, (2) \textbf{Robust Learning}: RLDA~\cite{Khodabandeh_2019_ICCV}, (3) \textbf{Feature-level adaptation}: DAF~\cite{Chen_2018_CVPR}, SCDA~\cite{zhu2019adapting}, MAF~\cite{He_2019_ICCV}, SWDA~\cite{Saito_2019_CVPR} and MDA~\cite{Xie_2019_ICCV_Workshops}, and (4) \textbf{Pixel-level adaptation + Feature-level adaptation}: DD-MRL~\cite{Kim_2019_CVPR} and PDA~\cite{Hsu_2019_CVPR_Workshops}. Moreover, we also provide ablation studies to validate the effectiveness of each module. Our baseline method is referred as 3DC, which is the Faster R-CNN model integrated with three domain classifiers. We alternately remove the ART and PSA module from the entire framework and report the performance. Note that removing the ART means we only remove the attention map while domain classifiers are still kept.

\vspace{-5mm}
\paragraph{\textit{Normal-to-Foggy}.} As shown in Table~\ref{table:city-foggy}, we achieve an mAP of 38.6\% on the weather transfer task, which is the best result among all the counterparts. Since detection performance before adaptation is different for each method, we point out that ``Gain'' is also a key criterion for fair comparison, which is ignored by previous work. In particular, we achieve a remarkable increase of +17.8\% over the source only model. Among all the feature-level adaptation methods, we improve the mAP by +2.6\% compared to MDA~\cite{Xie_2019_ICCV_Workshops}. Although we do not leverage extra pixel-level adaptation, our method still outperforms previous state-of-the-art PDA~\cite{Hsu_2019_CVPR_Workshops} by +1.7\%. Besides, with the help of coarse-to-fine feature adaptation on foregrounds, the proposed method brings improvements on all the categories than the 3DC model does, which shows that feature alignment on foregrounds can boost performance. Additionally, we find that the proposed method is comparable to or even better than the oracle model in several categories. It suggests that the performance which is similar to that of supervised learning methods can be achieved, if we effectively transfer knowledge across domains.

\vspace{-5mm}
\paragraph{\textit{Synthetic-to-Real}.} Table~\ref{table:sim-city} displays the results on the Synthetic-to-Real task. We obtain an average precision of 43.8\% on \textit{Car} and find that there is a slight gain of +0.8\% compared to SCDA~\cite{zhu2019adapting}. The reason is that knowledge transfer is much easier for single category, and many other methods can also adapt well. Further, one may wonder why the PSA module is still effective for single category adaptation, and we argue that it serves as another attention mechanism that focuses on foreground regions, which conveys some complementary clues to the ART module in this case.

\vspace{-5mm}
\paragraph{\textit{Cross-Camera}.} In Table~\ref{table:city-kitti}, we illustrate the performance comparison on the cross-camera task. The proposed method reaches an mAP of 41.0\% with a gain of +7.6\% over the non-adaptive model. Due to the fact that  scenes are similar across domains and the \textit{Car} sample dominate the two datasets, we can observe that the score on \textit{Car} is already good for the source only model. Compared with DAF~\cite{Chen_2018_CVPR} and MDA~\cite{Xie_2019_ICCV_Workshops}, our method reduces the influence of negative transfer in \textit{Car} detection. Meanwhile, our method also outperforms the baseline model (3DC) in the rest categories.

\begin{table}[t]
    \scriptsize
    \begin{center}
    \setlength{\tabcolsep}{2.5pt}
    \begin{tabular}{c|c|ccccc|c|c|c} 
    \toprule
    \multicolumn{10}{c}{\textbf{Cityscapes $\rightarrow$ KITTI}} \\
    \midrule 
        Method & Arch. & Person & Rider & Car & Truck & Train & \textbf{mAP} & \textbf{mAP*} & \textbf{Gain} \\
    \midrule \midrule
    DAF \cite{Chen_2018_CVPR} & V & 40.9 & 16.1 & 70.3 & 23.6 & 21.2 & 34.4 & 34.0 & 0.4 \\
    MDA \cite{Xie_2019_ICCV_Workshops} & V & \textbf{53.3} & 24.5 & 72.2 & 28.7 & \textbf{25.3} & \underline{40.7} & 34.0 & \underline{6.7} \\
    \midrule
    Source Only & V & 48.1 & 23.2 & \textbf{74.3} & 12.2 & 9.2 & 33.4 & - & - \\
    3DC (Baseline) & V & 45.8 & 27.0 & \underline{73.9} & 26.4 & 18.4 & 38.3 & 33.4 & 4.9 \\
    Ours w/o ART & V & 50.2 & 27.3 & 73.2 & \underline{29.5} & 17.1 & 39.5 & 33.4 & 6.1 \\
    Ours w/o PSA & V & \underline{50.5} & \underline{27.8} & 73.3 & 26.8 & 20.5 & 39.8 & 33.4 & 6.4 \\
    Ours & V & 50.4 & \textbf{29.7} & 73.6 & \textbf{29.7} & \underline{21.6} & \textbf{41.0} & 33.4 & \textbf{7.6} \\
    \midrule \midrule
    Oracle & V & 71.1 & 86.6 & 88.4 & 90.7 & 90.1 & 85.4 & - & - \\
    \bottomrule
    \end{tabular}
    \end{center}
    \caption{
        Results (\%) of the Cross-Camera adaptation scenario.
    }
    \label{table:city-kitti}
    \vspace{-5mm}
\end{table}

\begin{figure*}[t]
	\centering
	\footnotesize
   \begin{tabular}{c@{\hskip2pt}c@{\hskip2pt}c@{\hskip2pt}c}
      \includegraphics[width=4.2cm]{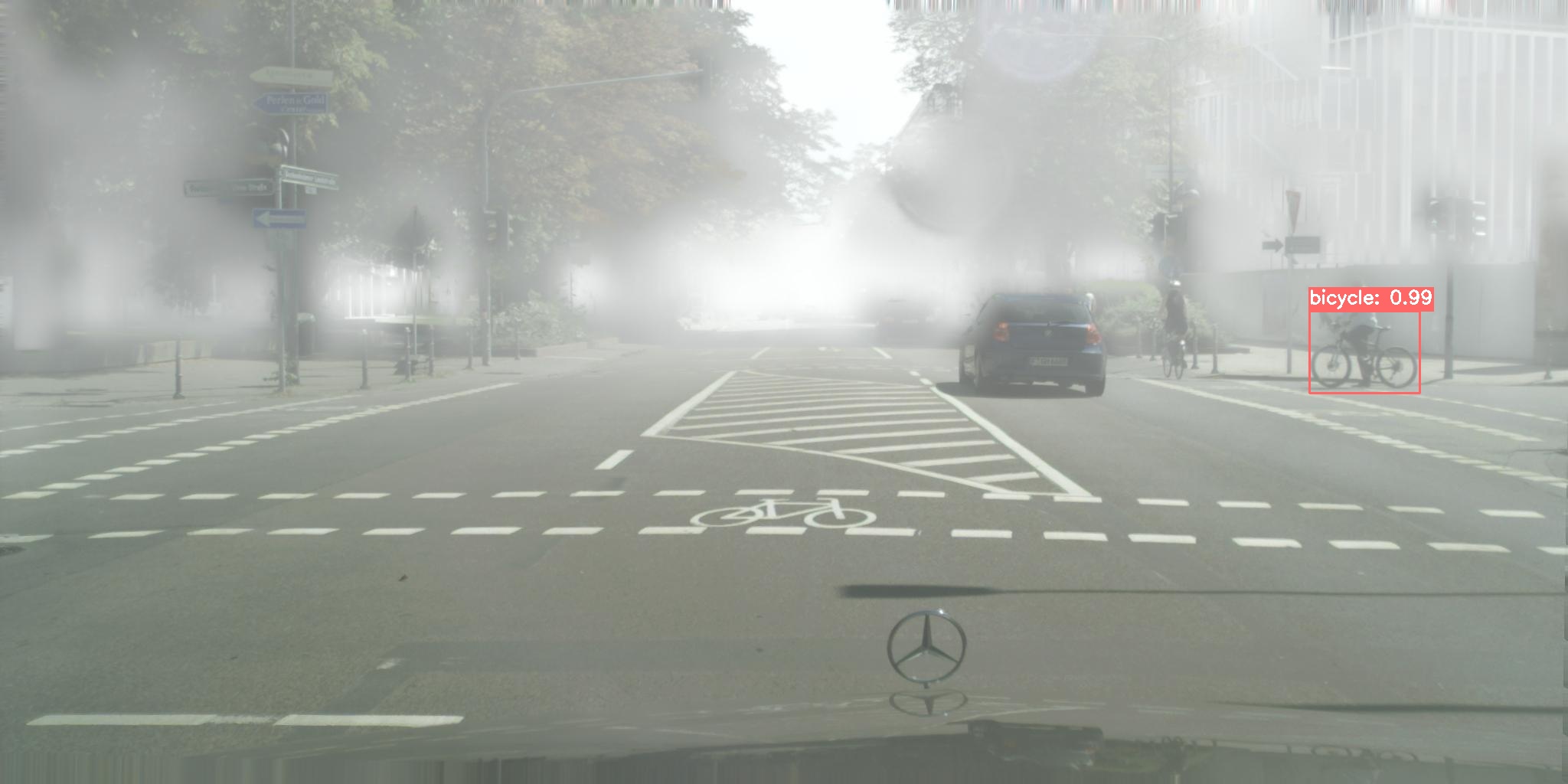}&%
		\includegraphics[width=4.2cm]{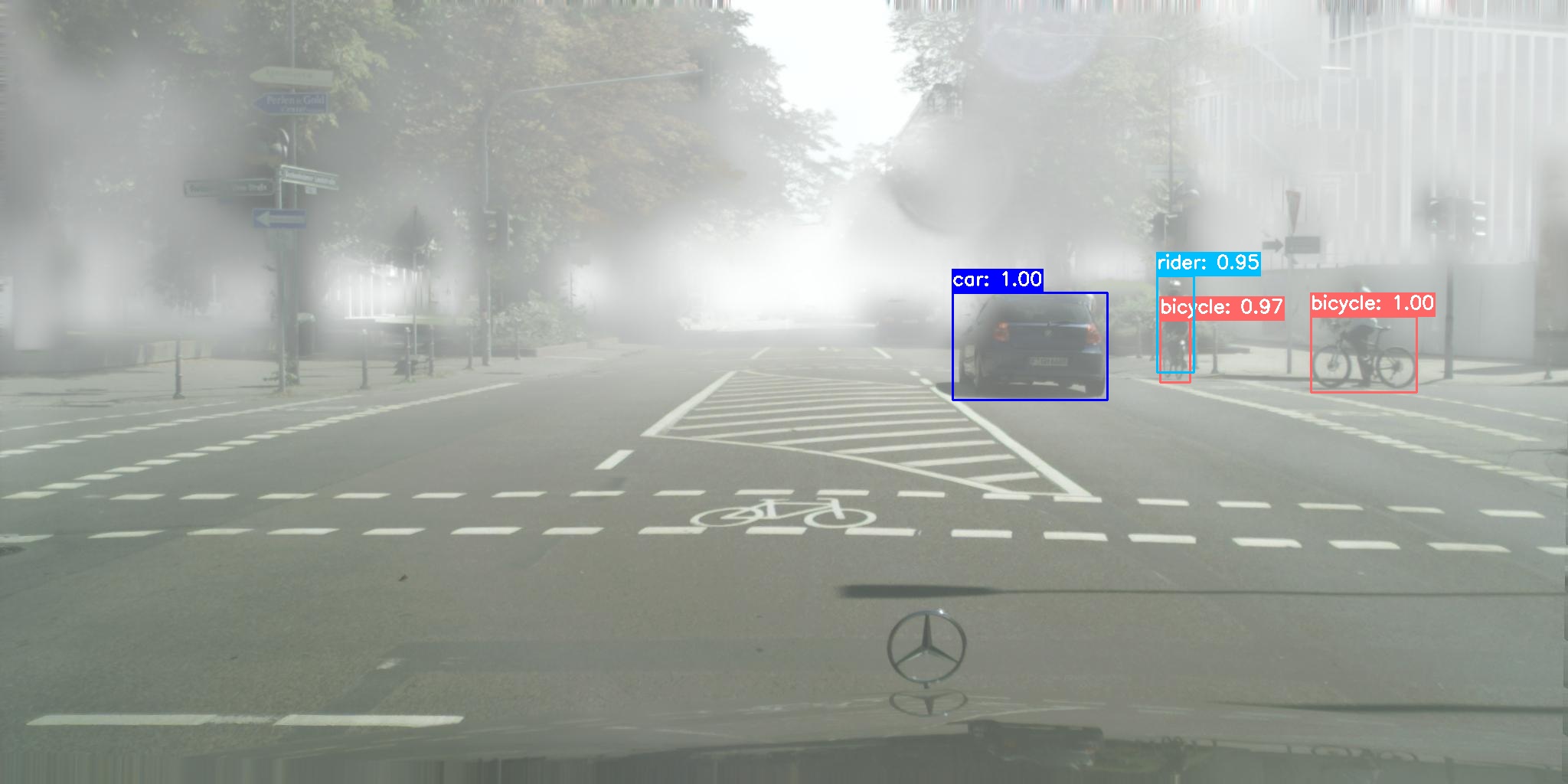}&%
		\includegraphics[width=4.2cm]{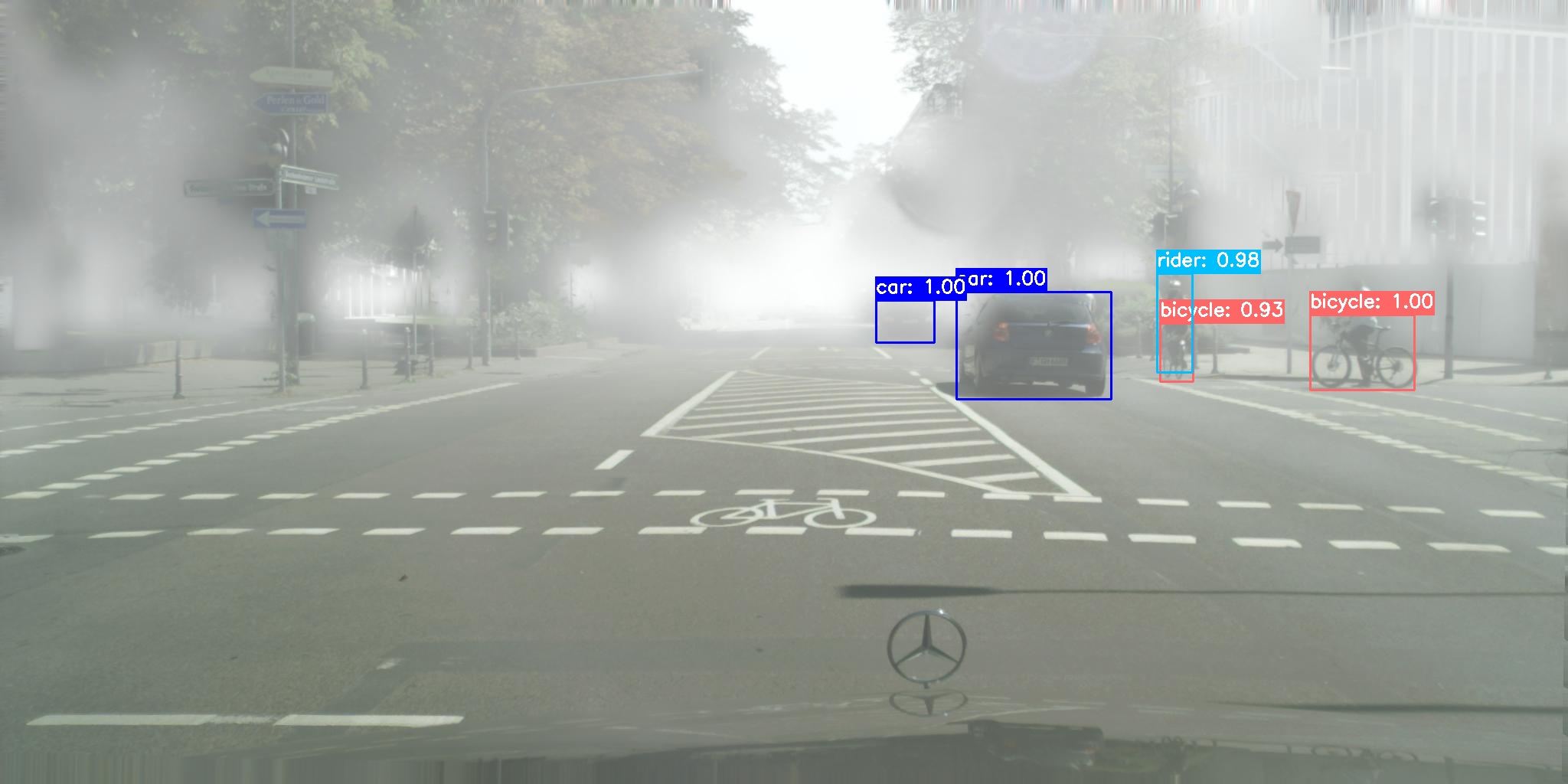}&%
      \includegraphics[width=4.2cm]{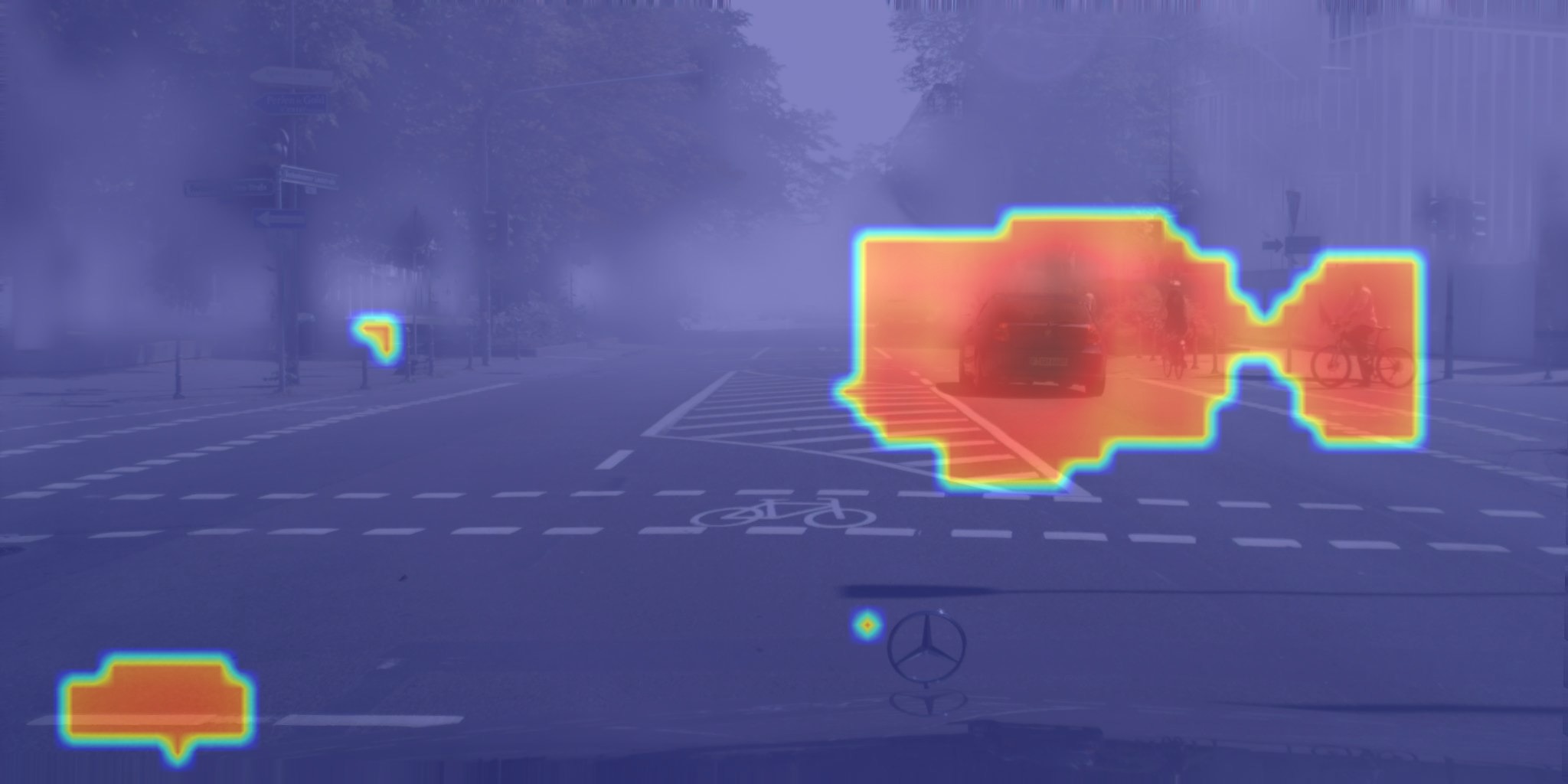}\\
		\includegraphics[width=4.2cm]{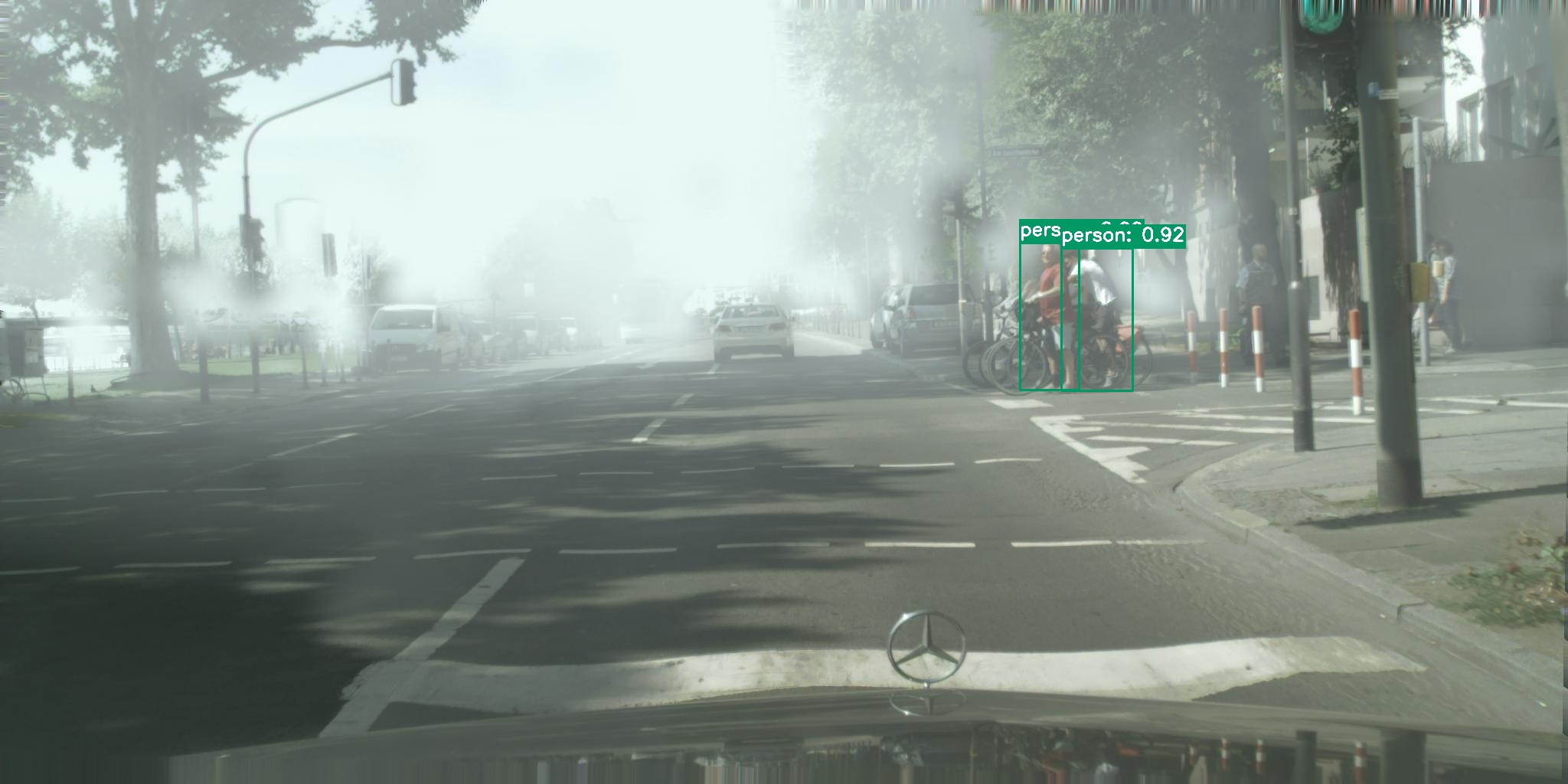}&%
		\includegraphics[width=4.2cm]{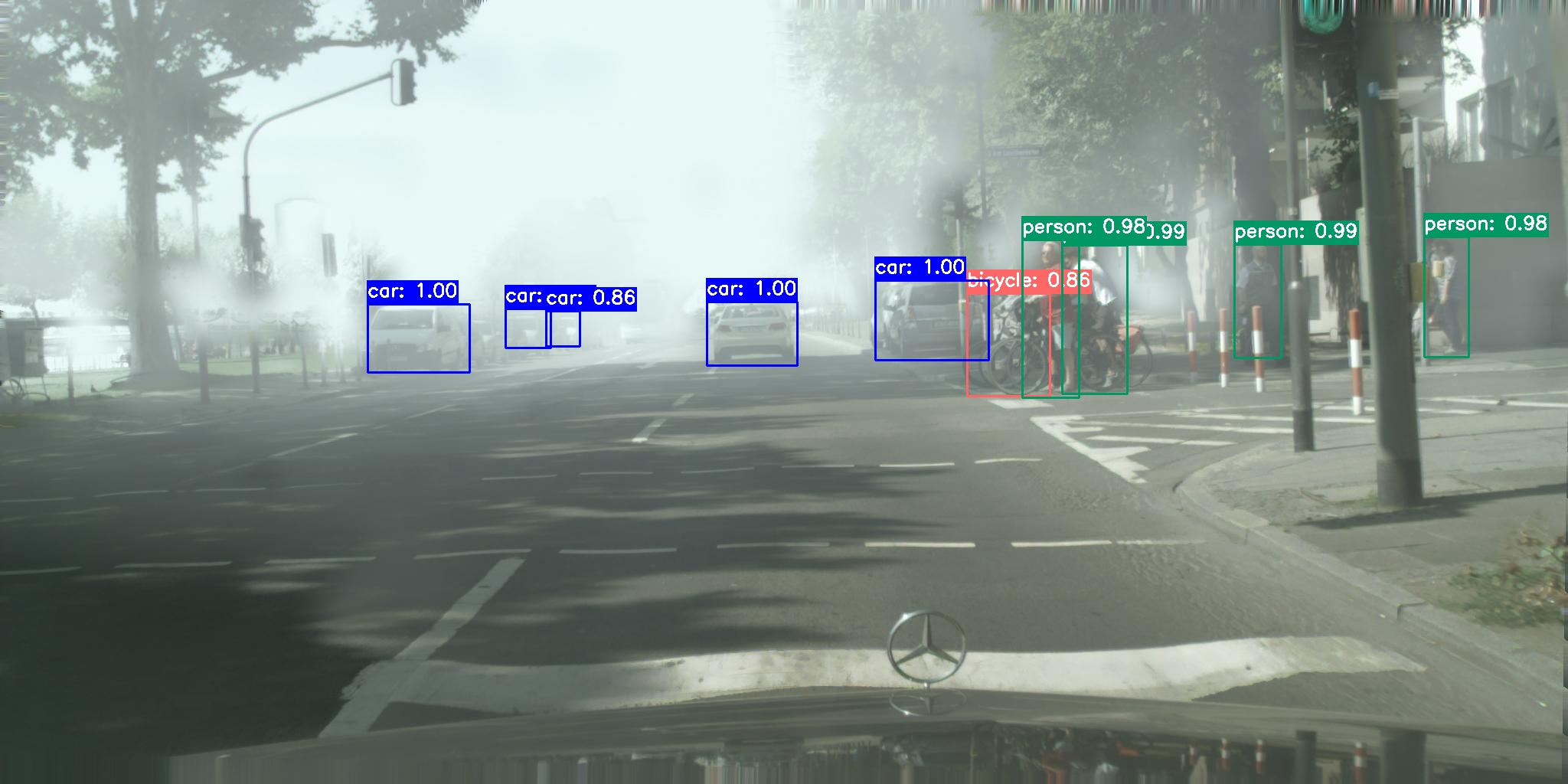}&%
		\includegraphics[width=4.2cm]{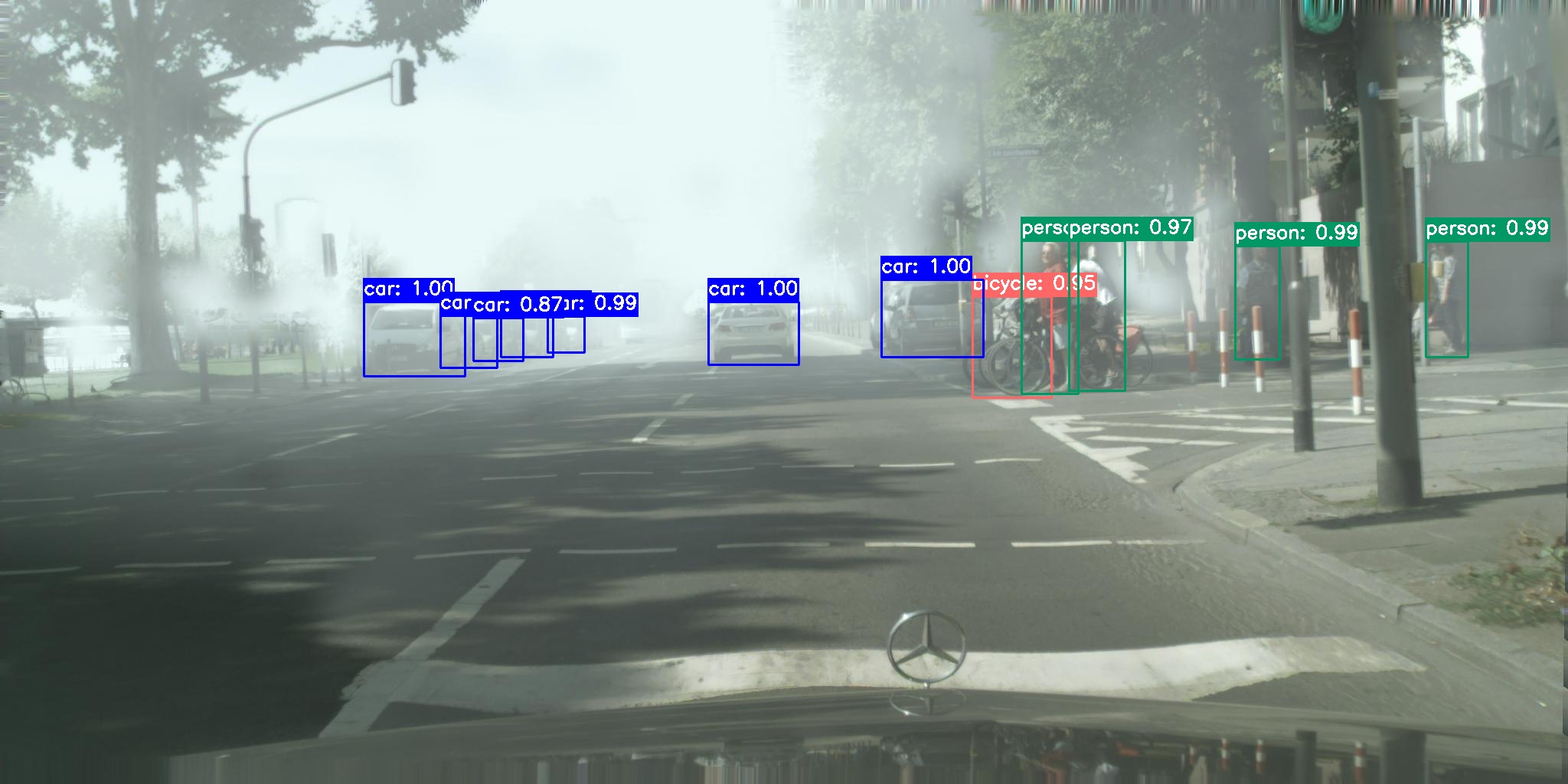}&%
      \includegraphics[width=4.2cm]{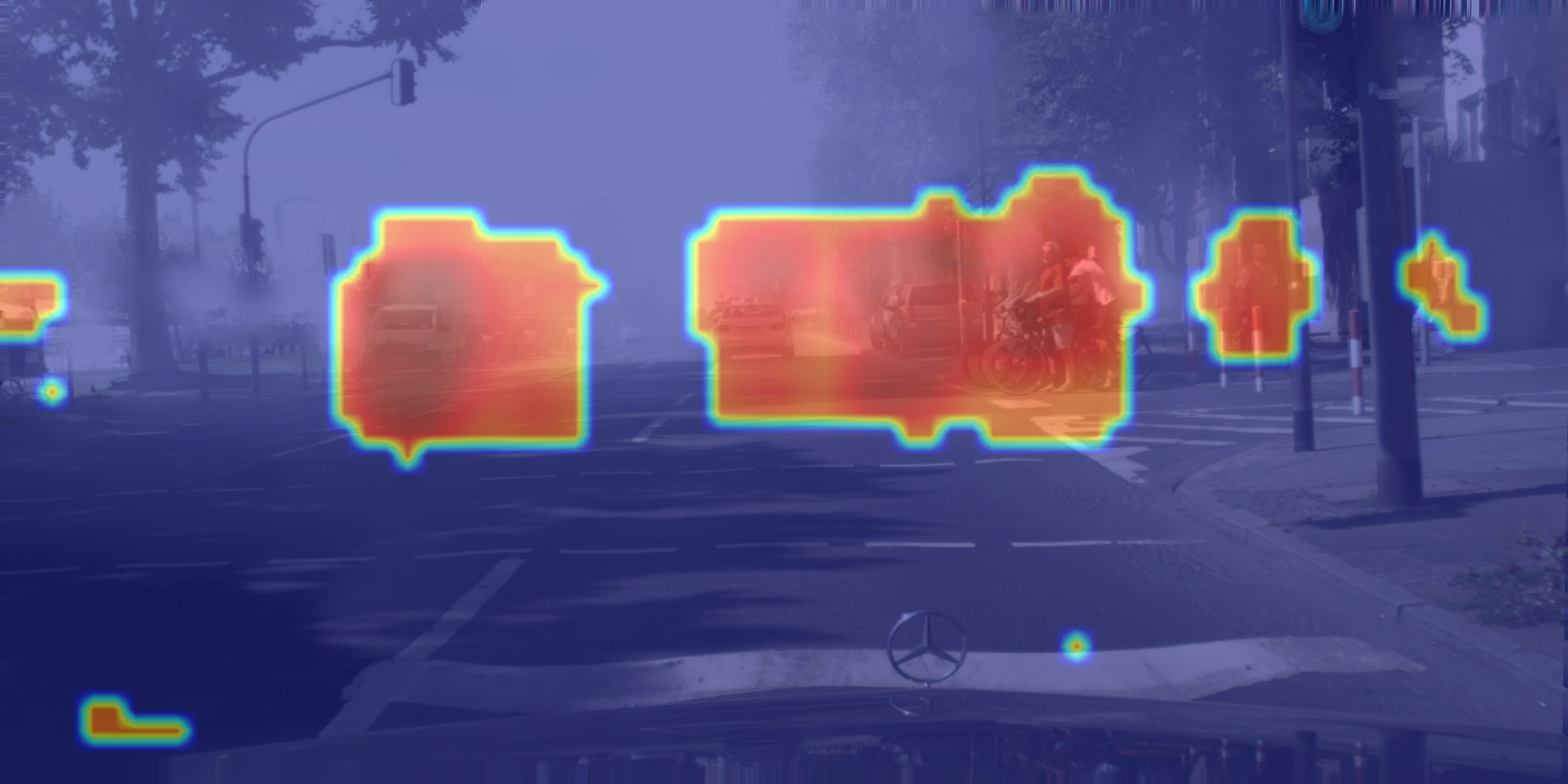}\\
      \includegraphics[width=4.2cm]{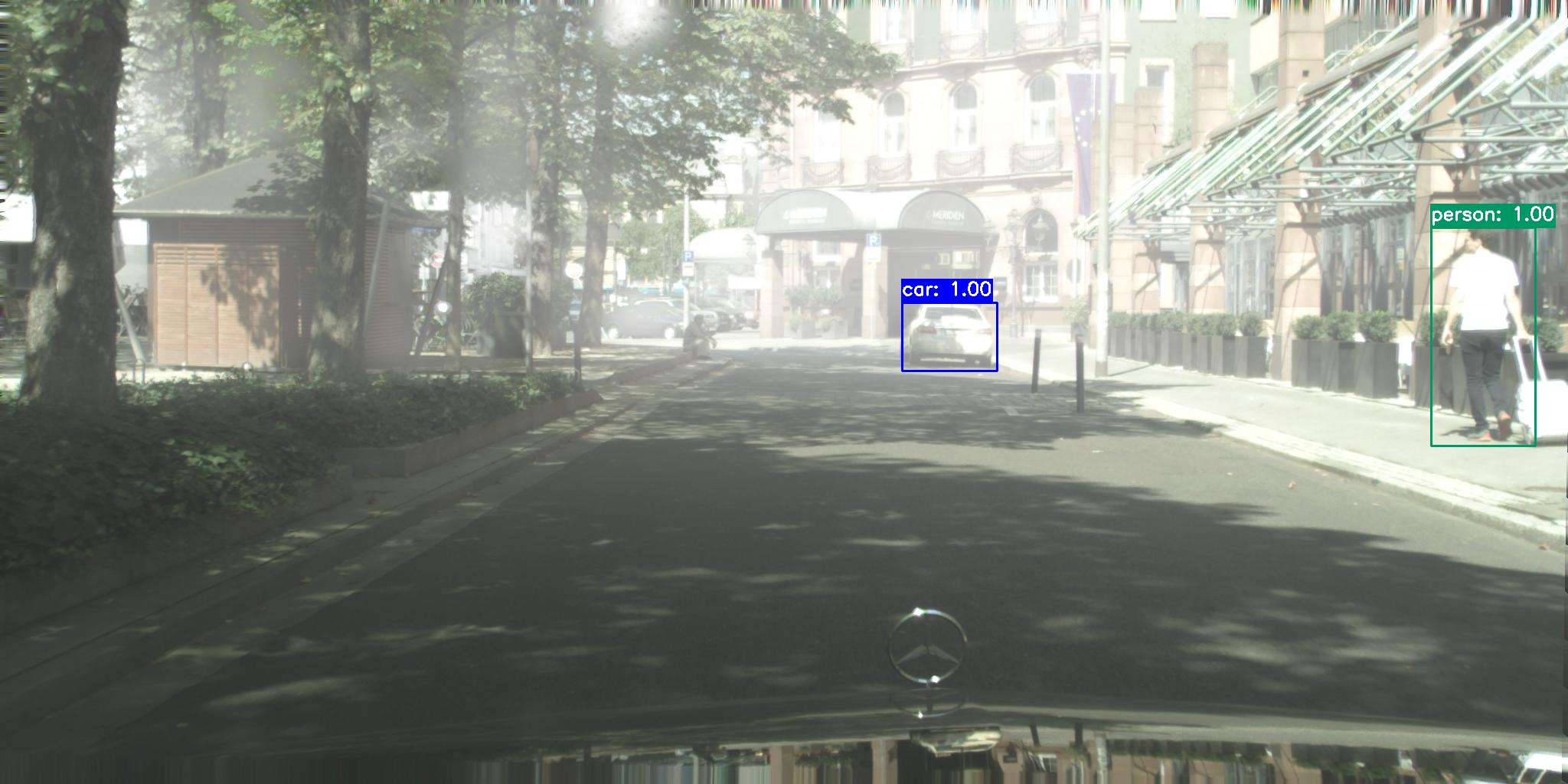}&%
		\includegraphics[width=4.2cm]{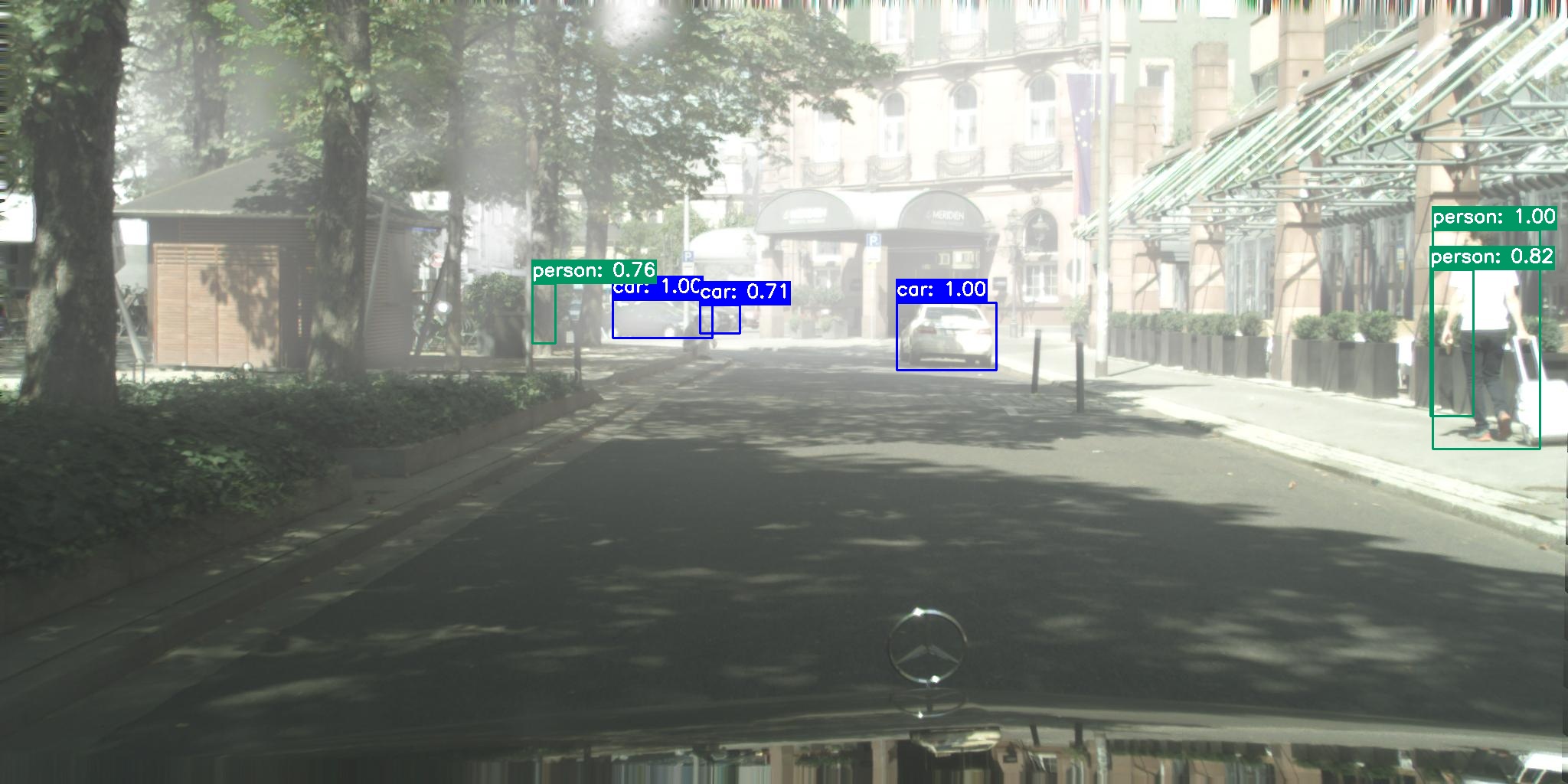}&%
		\includegraphics[width=4.2cm]{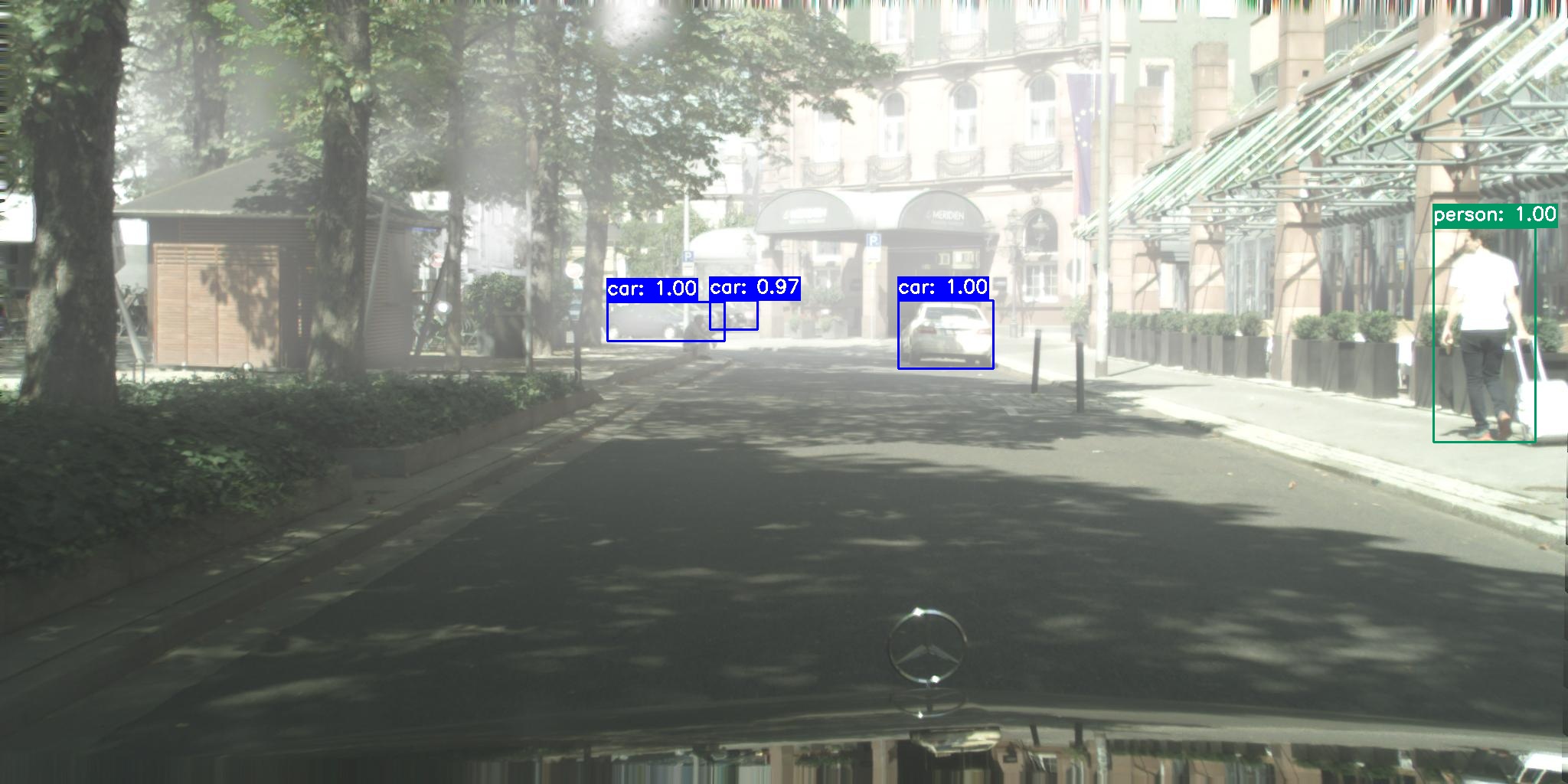}&%
      \includegraphics[width=4.2cm]{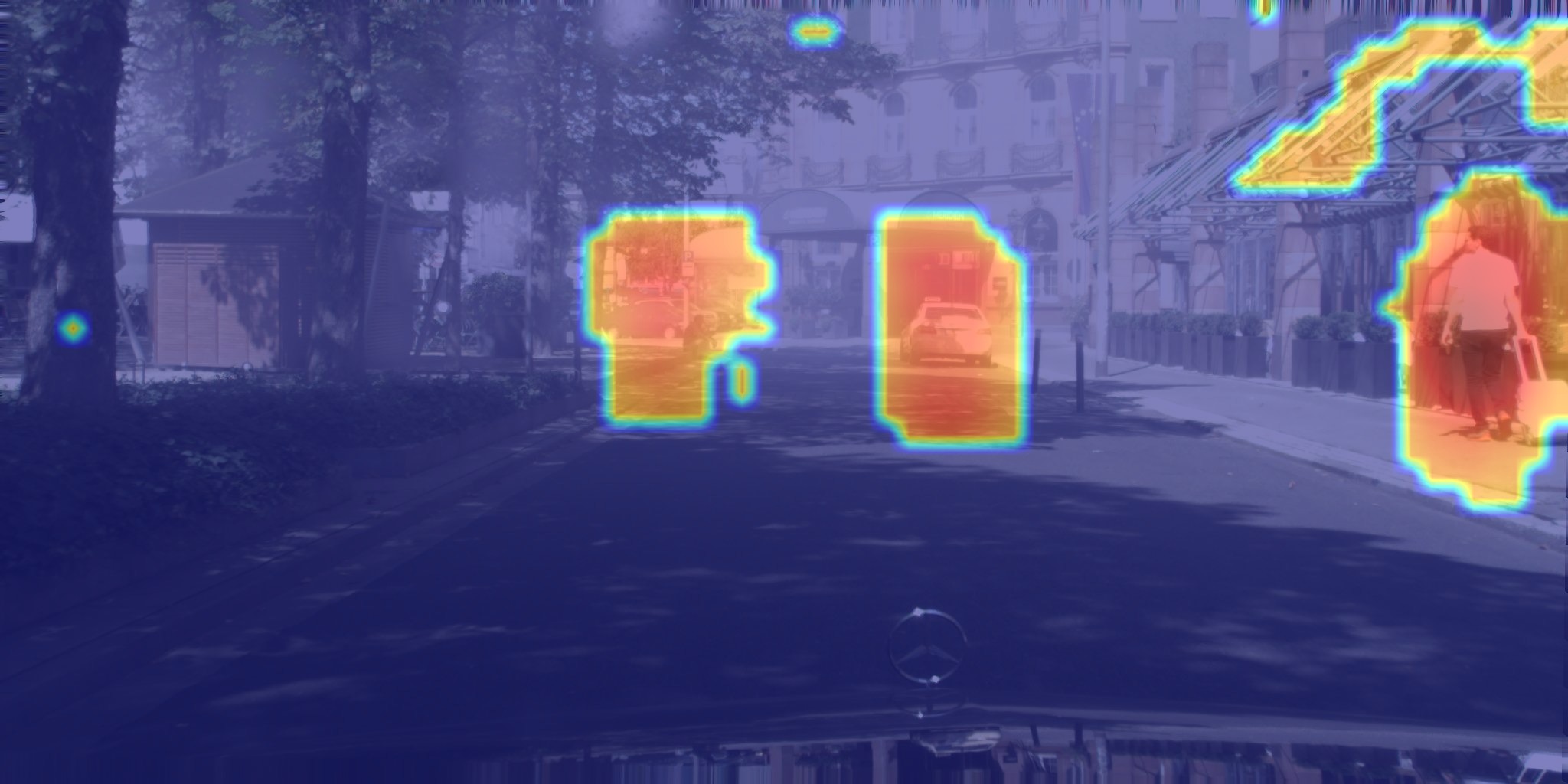}\\
		(a) Source Only & (b) SWDA & (c) Ours & (d) Attention Map \\
	\end{tabular}\vspace{2mm}
	\caption{Qualitative results on the \textit{Normal-to-Foggy} adaptation scenario. (a)-(c): The detection results of the Source Only model, SWDA and the proposed method. (d): Visualization of the corresponding attention maps (best viewed by zooming in).}\label{fig:results}
	\vspace{-3mm}
\end{figure*}

\subsection{Further Analysis}
\vspace{-1mm}

\begin{figure}
   \centering
   \includegraphics[width=1.0\linewidth]{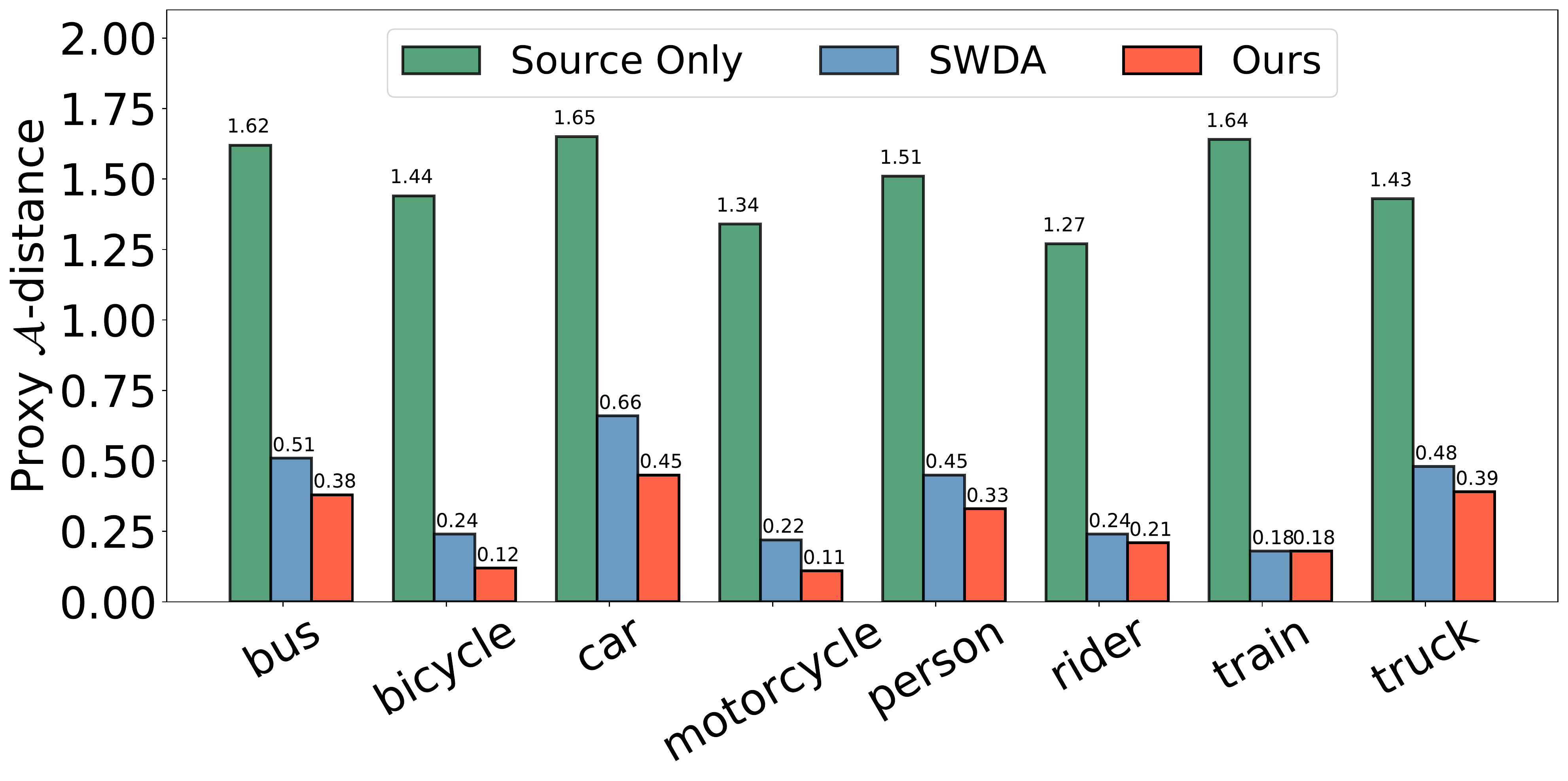}
   \caption{Feature distribution discrepancy of foregrounds.}
   \label{fig:distance} 
   \vspace{-3mm}
\end{figure}

\paragraph{Feature distribution discrepancy of foregrounds.} The theoretical result in~\cite{ben2007analysis} suggests that $\mathcal{A}$-distance can be used as a metric of domain discrepancy. In practice, we calculate the Proxy $\mathcal{A}$-distance to approximate it, which is defined as $d_{\mathcal{A}} = 2(1 - \epsilon)$. $\epsilon$ is the generalization error of a binary classifier (linear SVM in our experiments) that tries to distinguish which domain the input features come from. Figure~\ref{fig:distance} displays the distances for each category on the \textit{Normal-to-Foggy} task with the features of ground truth foregrounds extracted from the models of \textit{Source Only}, \textit{SWDA} and \textit{Ours}. Compared with the non-adaptive model, \textit{SWDA} and \textit{Ours} reduce the distances in all the categories by large margins, which demonstrates the necessity of domain adaptation. Besides, since we explicitly optimize the prototypes of each category by PSA, we achieve a smaller feature distribution discrepancy of foregrounds than the others do.

\begin{figure}
   \centering
   \includegraphics[width=1.0\linewidth]{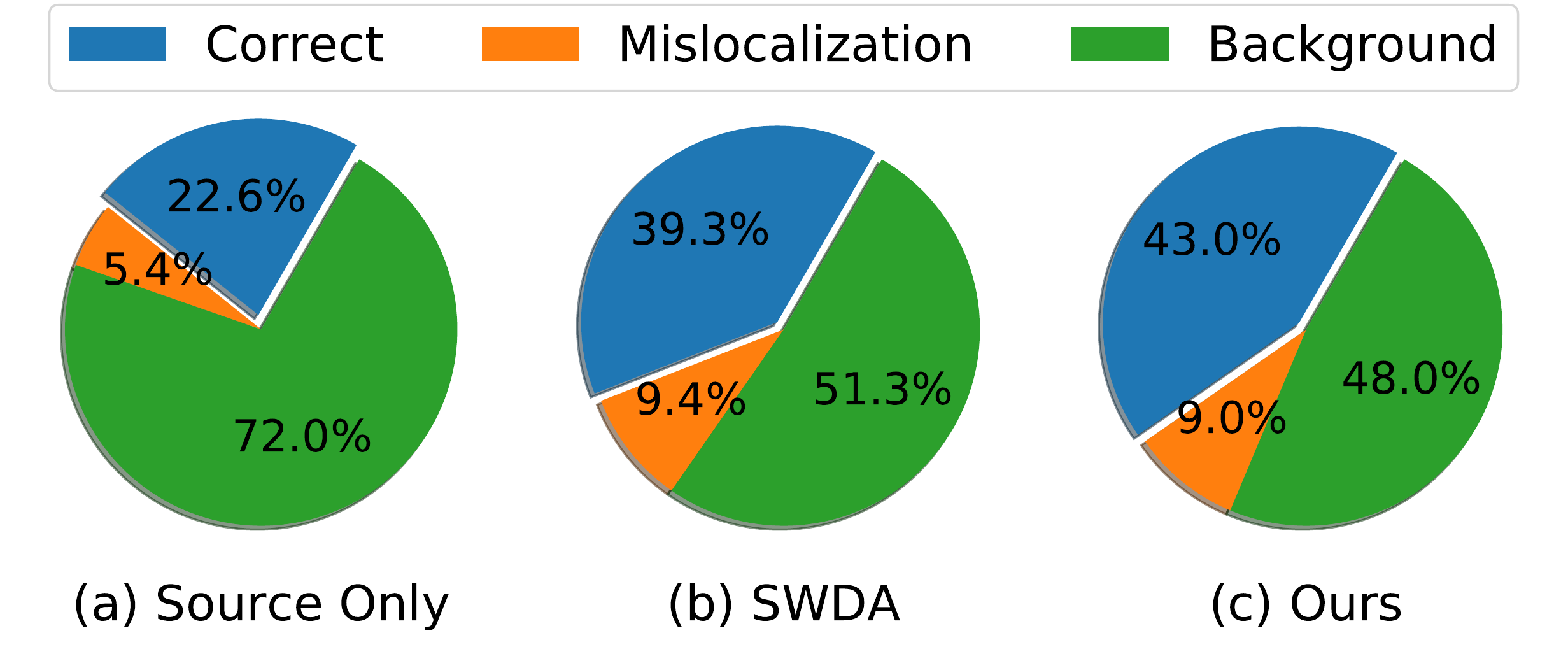}
   \caption{Error analysis of highest confident detections.}
   \label{fig:error} 
   \vspace{-3mm}
\end{figure}

\vspace{-4mm}
\paragraph{Error analysis of highest confident detections.} To further validate the effect of the proposed framework for cross-domain object detection, we analyze the errors of the models of \textit{Source Only}, \textit{SWDA} and \textit{Ours} caused by highest confident detections on the \textit{Normal-to-Foggy} task. We follow~\cite{hoiem2012diagnosing, Chen_2018_CVPR, Cai_2019_CVPR} to categorize the detections into three error types: \textbf{1) Correct} (IoU with GT $\ge$ 0.5), \textbf{2) Mislocalization} (0.3 $\le$ IoU with GT $<$ 0.5), and \textbf{3) Background} (IoU with GT $<$ 0.3). For each category, we select top-$K$ predictions to analyze the error type, where $K$ is the number of ground truths in this category. We report the mean percentage of each type across all categories in Figure~\ref{fig:error}. We can see that the \textit{Source Only} model seems to take most of backgrounds as false positives (green color). Compared with \textit{SWDA}, we improve the percentage of correct detections (blue color) from 39.3\% to 43.0\% and reduce other error types simultaneously. The results indicate that the proposed framework can effectively increase true positives and reduce false positives, resulting in better detection performance.

\vspace{-5mm}
\paragraph{Qualitative results.} Figure~\ref{fig:results} shows some qualitative results. Due to the domain shift problem, the \textit{Source Only} model simply detects some salient objects as shown in (a).  From (b) to (c), we can observe that the proposed method not only increases true positives (detects more cars in the first and second row), but also reduces false positives (discards persons in the third row), which is consistent with previous analysis. Further, we visualize the attention maps generated from the ART module. Despite some noise, the attention maps well locate the foreground regions, which is beneficial to knowledge transfer across domains.

\vspace{-3mm}
\section{Conclusion}
\vspace{-1mm}
In this paper, we present a novel coarse-to-fine feature adaptation approach to address the issue of cross-domain object detection. The proposed framework achieves the goal with the incorporation of two delicately designed modules, \ie, ART and PSA. The former highlights the importance of the foreground regions figured out by the attention mechanism in a category-agnostic way, and aligns their feature distributions across domains. The latter takes the advantage of prototypes to perform fine-grained adaptation of foregrounds at the semantic level. Comprehensive experiments are conducted on various adaptation scenarios and state-of-the-art results are reached, demonstrating the effectiveness of the proposed approach.


\vspace{-4mm}
\paragraph{Acknowledgment.} This work is funded by the National Key Research and Development Plan of China under Grant 2018AAA0102301, the Research Program of State Key Laboratory of Software Development Environment (SKLSDE-2019ZX-03), and the Fundamental Research Funds for the Central Universities.

{\small
\bibliographystyle{ieee_fullname}
\bibliography{egbib}
}

\end{document}